%% file: Formatting_Instructions_For_NeurIPS_2026/neurips_2026.tex
\documentclass{article}

 \usepackage[main, final]{neurips_2026}

\usepackage{amsmath,amssymb,amsfonts}
\usepackage{algorithmic}
\usepackage{algorithm}
\usepackage{graphicx}
\usepackage{wrapfig}
\usepackage{textcomp}
\usepackage{xcolor}
\usepackage{multirow}
\usepackage{booktabs}
\usepackage{tabularx}
\usepackage{makecell}  
\usepackage{pifont}
\usepackage{hyperref}  
\usepackage{xcolor}  
\usepackage{caption}
\usepackage{subcaption}  
\usepackage{lipsum}
\usepackage{etoc} 

\hypersetup{
  colorlinks   = true,    
  urlcolor     = blue,    
  linkcolor    = blue,    
  citecolor    = blue      
}
\usepackage[utf8]{inputenc} 
\usepackage[T1]{fontenc}    
\usepackage{url}            
\usepackage{booktabs}       
\usepackage{amsfonts}       
\usepackage{nicefrac}       
\usepackage{microtype}      
\usepackage{xcolor}         

\newcommand{\eg}{\textit{e.g.}}
\newcommand{\ie}{\textit{i.e.}}
\newcommand{\name}{TRACE}
\title{TRACE: Structure-Aware Character Encoding for Robust and Generalizable Document Watermarking}

\author{%
    \textbf{Jiale Meng}$^{1}$ \quad
    \textbf{Jie Zhang}$^{2}$ \quad
    \textbf{Runyi Hu}$^{1}$ \\
    \textbf{Zheming Lu}$^{3}$\thanks{Corresponding author.} \quad
    \textbf{Tianwei Zhang}$^{1}$ \quad
    \textbf{Yiming Li}$^{1}$ \\[0.6em]
    $^{1}$Nanyang Technological University \quad
    $^{2}$SiliconProspect AI \\
    $^{3}$Zhejiang University \\[0.4em]
    {\tt\small jiale.meng@ntu.edu.sg \quad
    zheminglu@zju.edu.cn} \\
    {\tt\small liyiming.tech@gmail.com}
}

\begin{document}

\maketitle

\begin{abstract}
  We propose TRACE, a structure-aware framework leveraging diffusion models for localized character encoding to embed data. Unlike existing methods that rely on edge features or pre-defined codebooks, \name{} exploits character structures that provide inherent resistance to noise interference due to their stability and unified representation across diverse characters. Our framework comprises three key components: \textbf{(1)} adaptive diffusion initialization that automatically identifies handle points, target points, and editing regions through specialized algorithms including movement probability estimator (MPE), target point estimation (TPE) and mask drawing model (MDM), \textbf{(2)} guided diffusion encoding for precise movement of the selected point, and \textbf{(3)} masked region replacement with a specialized loss function to minimize feature alterations after the diffusion process. Comprehensive experiments demonstrate \name{}'s superior performance over state-of-the-art methods, achieving more than 5 dB improvement in PSNR and 5\% higher extraction accuracy following cross-media transmission. \name{} achieves broad generalizability across multiple languages and fonts, making it particularly suitable for practical document security applications. Code is available at \href{https://github.com/ELLA-MENG/TRACE}{https://github.com/ELLA-MENG/TRACE}.
\end{abstract}

\section{Introduction}
\label{sec:intro}

Documents are ubiquitous in daily life, appearing as official papers, printed materials, and digital documents. Tracing their origins is critical for protecting textual content and preventing misuse. While numerous NLP-based watermarking techniques \cite{abdelnabi2021adversarial, zhang2024remark} trace the provenance of LLM-generated text, they are designed for purely digital text and thus fail in many real-world settings. For example, they cannot protect visualized text, such as printed, handwritten, stylized, or stamped documents, nor cases where the textual content cannot be modified due to sensitivity or authenticity constraints.

\begin{figure}[!t]
\centering
\includegraphics[width=0.95\textwidth]{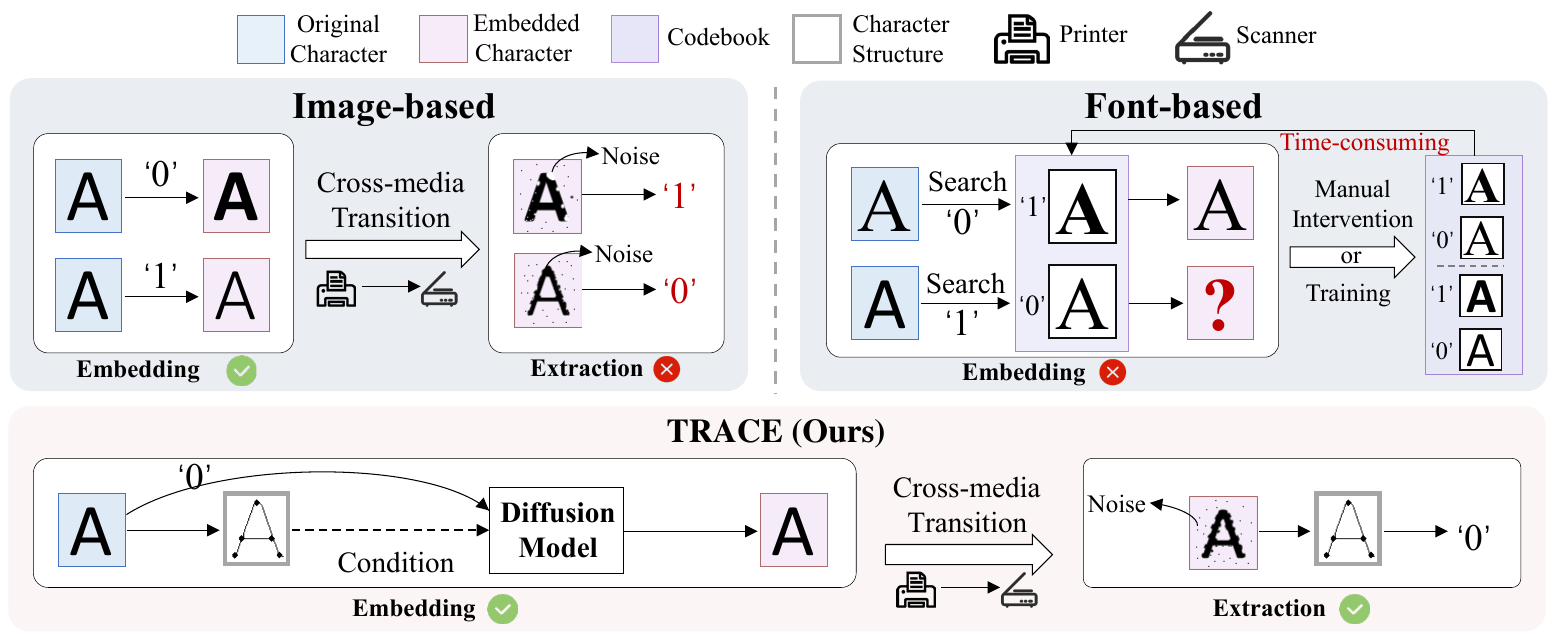}
\caption{\textbf{Illustration of different document hiding methods.} Image-based methods embed information by adjusting the proportion of black pixels but are vulnerable to noise from cross-media transmission, leading to extraction errors. Font-based methods rely on predefined character codebooks, which limits their generalizability when encountering unseen characters. In contrast, our method leverages character structures to guide diffusion-based encoding, ensuring broad applicability. Moreover, character structures are more stable to noise, providing strong robustness.}
\label{fig:intro}
\vspace{-1.5em}
\end{figure}

Although many data hiding methods have been developed for color images \cite{hu2025mask, sander2025watermark,huang2024robust}, they cannot be directly applied to documents, which are typically composed of characters on smooth backgrounds rather than complex colors and textures. To address this gap, document-oriented hiding methods have been proposed, mainly including image-based \cite{wu2004data, meng2025coremark} and font-based \cite{yang2023autostegafont, xiao2018fontcode} approaches. 
As shown in Fig.~\ref{fig:intro}, image-based methods treat a document as an image and embed data by altering the black-white pixel ratio of each character. However, such ratios are easily disrupted by cross-media noise (\eg, printing or scanning), making these methods less robust. Increasing the number of modified pixels can improve robustness, but at the cost of stealthiness. Font-based methods instead encode information by replacing characters with predefined variants from a codebook. While more robust to noise, their generalizability is inherently limited, as a codebook cannot cover all possible characters across diverse languages, fonts, handwritten styles, and artistic typography.
Overall, an effective document hiding method should simultaneously achieve robustness, generalizability, and invisibility. However, to the best of our knowledge, no prior method satisfies all three. This raises an important question: \textit{Can we design a document hiding method that offers strong robustness, broad generalizability, and high imperceptibility?}

The answer to the aforementioned question is positive, although its solution is non-trivial. In this paper, we argue that we can implant data based on character structure encoding to meet all aforementioned requirements. Intuitively, character structures are less susceptible to noise interference since it is more stable; the structure can be regarded as a skeleton consisting of keypoints and connecting lines (see Fig.~\ref{fig:intro}), thus providing a unified representation across diverse characters; it can also ensure stealthiness since modifying a few pixels around the structure will not change its appearance. 

To do so, we propose a structure-based glyph modification method for document hiding (dubbed `\name{}'), inspired by Dragdiffusion \cite{shi2024dragdiffusion}. In general, our \name{} exploits keypoints to guide diffusion models to generate encoded characters by moving specific keypoints (\ie, handle points) to their corresponding target positions. Specifically, the information embedding stage consists of three main steps: \textbf{(1)} adaptive diffusion initialization, \textbf{(2)} guided diffusion encoding (GDE), and \textbf{(3)} masked region replacement (MRR). For the first step, the key elements guiding the diffusion process are determined, including handle point, target point, and editing region, through a systematic process involving the motion probability estimator (MPE), target point estimation (TPE), and mask drawing module (MDM), respectively. The MPE adaptively computes a score for each keypoint and selects the point with the highest score as the handle point, where a higher score indicates greater motion probability. Subsequently, the TPE determines the target point based on the directional properties of the stroke containing the handle point, ensuring watermark embedding while preserving the structural consistency of the character. MDM adaptively defines the editing area based on the handle point and target point, significantly enhancing the precision of the embedding operation. Upon obtaining the key elements, GDE exploits DragDiffusion to generate the encoded character. In particular, to minimize feature alterations in non-edited regions caused by the diffusion process, we introduce a simple yet effective technique (\ie, MRR), which replaces the content in the masked region of the original image with corresponding content from the edited image. Concurrently, we formulate a specialized loss function to ensure feature consistency within the masked region before and after editing, thereby preserving the perceptual integrity of the encoded character while maintaining the validity of the embedded data.

In summary, our main contributions are three-fold. \textbf{(1)} We propose a novel paradigm by embedding information into character structures, achieving a better balance between robustness, generalizability, and imperceptibility. \textbf{(2)} We design a structure-guided hiding scheme and a simple yet effective method (\ie, `\name{}') by leveraging diffusion models. \textbf{(3)} Extensive experiments are conducted, which verify the robustness, generalizability, and invisibility of our method, and its resistance to potential attacks. We hope our method can provide new insights into document hiding, to facilitate trustworthy document sharing and leakage attribution.

\section{Related Work}
\label{sec:related work}

\subsection{Hiding Information in Documents}
\noindent\textbf{Font-based Methods} \cite{yang2023autostegafont, xiao2018fontcode, qi2019robust, yao2024embedding, sun2024invertedfontnet,he2024watermarking,wong2025fontguard}. Xiao et al. \cite{xiao2018fontcode} gathered volunteers to select the glyphs most similar to the original font from a set of glyph variants. This process was repeated for different fonts to construct a glyph codebook. \cite{qi2019robust} generated character variants by adjusting the position of character strokes. Both methods, however, rely heavily on manual participation to construct the codebook, which makes them impractical. 
To address this, Yang et al. \cite{yang2023autostegafont} made some improvements by proposing a two-stage learning framework that automatically generates encoded characters. Although this method eliminates the need for manual intervention in constructing the character codebook, it does not fundamentally address the issue of generalizability. For example, when encountering glyphs that have not been pre-trained, this method still cannot succeed. More recently, FontGuard~\cite{wong2025fontguard} embeds watermarks by manipulating learned font-style features, enabling generalization to unseen fonts without retraining.

\vspace{0.3em}
\noindent\textbf{Image-based Methods} \cite{wu2004data, yang2023language, tan2019print,huang2024robust,li2023robust, meng2025coremark}. Following Wu et al.'s \cite{wu2004data} foundational work on pixel flipping score calculation, many subsequent methods embed information by flipping pixels to establish a relationship between the embedded data and black pixel statistical features. However, these methods are sensitive to the noise introduced during cross-media transmission, which can disrupt pixel features and significantly degrade robustness.

\vspace{0.3em}
\noindent\textbf{Format-based Methods} \cite{brassil1995electronic, huang2001interword, zou2005formatted, brassil1999copyright}. These methods typically embed watermarks by adjusting the spacing between lines or words, which generally suffer from low embedding capacity and poor robustness \cite{brassil1999copyright}. Additionally, some approaches require the original document for data extraction, posing significant challenges for practical applications.

\vspace{0.3em}
\noindent\textbf{Linguistic-based Methods} \cite{kirchenbauer2023watermark, dathathri2024scalable, yang2022tracing, he2022cater}. This approach conceals data by modifying textual syntactic and semantic properties. However, such content alterations are generally prohibited for critical documents like contracts and confidential files, thereby limiting the generalizability of the method.

Beyond document-specific methods, recent neural watermarking studies integrate watermarks into models or generation processes for copyright protection~\cite{song2024protecting, fernandez2023stable}. Unlike these approaches, \name{} performs structure-aware watermark embedding by modifying character skeletons. Despite their different embedding mechanisms, these methods share the goal of protecting content ownership and tracing provenance, reflecting a broader shift toward model-aware and structure-aware watermarking. More discussions and comparisons are in Appendix~\ref{sec:appdixrelatedwork}.

\subsection{Diffusion-based Image Editing}
Diffusion models \cite{ho2020denoising, nichol2021improved} have garnered significant attention for their high-quality generation capabilities. Research has leveraged these models for image editing through prompt-based \cite{ling2021editgan, Zhang2023CVPR, zhao2024ultraedit} and point-based approaches \cite{shi2024dragdiffusion, liu2024drag, shin2024instantdrag}. Prompt-based methods enable image manipulation via textual prompt modifications but typically lack spatial precision. Conversely, point-based techniques facilitate targeted adjustments in specific regions via point-to-point transformations. To the best of our knowledge, no prior work has explored diffusion-based image editing for watermarking applications.

\section{Methodology}
\label{sec:Methodology}

\begin{figure*}[t]
	\centering
	\includegraphics[width=0.95\textwidth]{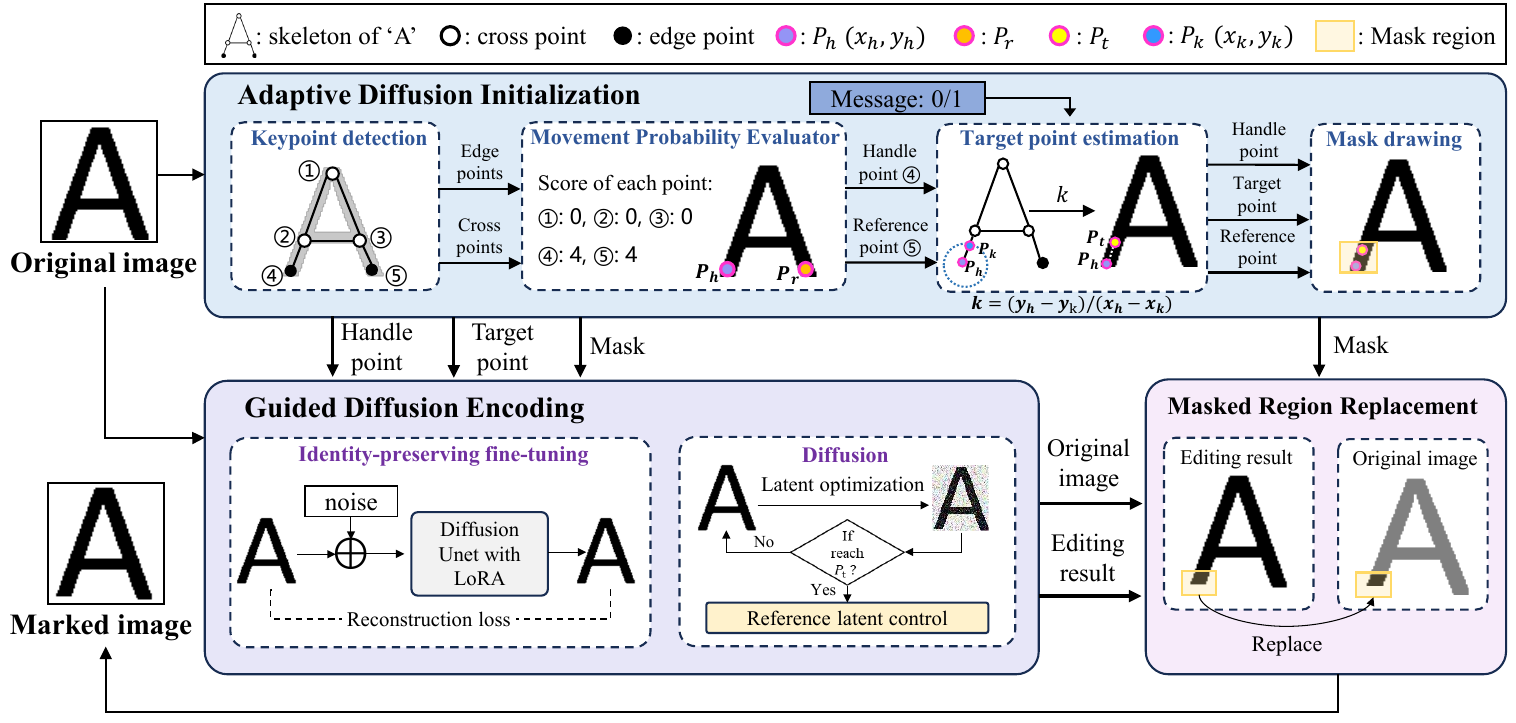}
    \vspace{-0.5em}
	\caption{\name{} Embedding Pipeline. \textbf{Step 1. Adaptive Diffusion Initialization:} Handle points, target points, and masks are identified to guide the subsequent diffusion process. \textbf{Step 2. Guided Diffusion Encoding:} The diffusion model's UNet is fine-tuned via LoRA to better capture original image features. The marked image is then generated by controlled movement of handle points to target points within the predefined mask region. \textbf{Step 3. Masked Region Replacement:} The content within the mask of the original image is replaced with the corresponding encoded segments.}
    \label{fig:pipeline}
	\vspace{-1em}
\end{figure*}

In general, our \name{} has two main stages: data embedding and data extraction. The embedding stage consists of three steps: \textbf{(1)} adaptive diffusion initialization, \textbf{(2)} guided diffusion encoding, and \textbf{(3)} masked region replacement. Fig.~\ref{fig:pipeline} shows the embedding framework of \name{}.

\subsection{Adaptive Diffusion Initialization}
\label{sec-adi}
As shown in the blue area of Fig.~\ref{fig:pipeline}, in this stage, given a text image $ I_{\text{cover}}$, we first extract endpoints and junction points from its skeleton $ I_{\text{ske}} $, which are then fed into the movement probability evaluator (MPE) to select the handle point $ P_h $ and its reference point $ P_r $. Subsequently, the MPE outputs are used to calculate the target point $ P_t $ according to the message.  Finally, both $ P_h $ and  $ P_t $ are processed by the mask drawing module (MDM) to generate the mask $ \mathcal{M} $.

\vspace{0.3em}
\noindent \textbf{Keypoint Detection.}
As shown in Fig.~\ref{fig:pipeline}, to guide the diffusion-based editing process, we first extract character keypoints, which are categorized into two distinct types: endpoints and junction points, denoted by sets $E$ and $C$, respectively. Our keypoint extraction network builds upon the lightweight OpenPose architecture \cite{cao2019openpose}, by selectively retaining only the keypoint computation components while eliminating elements related to keypoint matching and grouping. Moreover, we modify the output confidence map to comprise three channels: an endpoint heatmap, a junction point heatmap, and a background heatmap. Note that the keypoint extraction methodology is flexible and user-defined, allowing seamless integration of improved detection techniques.

\vspace{0.3em}
\noindent \textbf{Movement Probability Evaluator.}
\label{sec:MPE}
After identifying junction points and endpoints, we select the optimal point to serve as the handle point for manipulation. The movement probability evaluator (MPE) automates this selection process for each unique character, which enhances extraction efficiency while maintaining the visual consistency of characters before and after encoding.

\noindent\textit{Reference-point Candidate Construction.}
Before delving into the operating mechanism of MPE, we first introduce a critical component: the reference point $ P_r $. MPE utilizes $ P_r $ as a reference to determine the handle point $ P_h $.
For each point $ p_i^e $ in set $ E $ with coordinates $ \left(x_i^e, y_i^e\right) $, the reference point $ p_{i,j}^r$ with coordinates $\left(x_{i,j}^r, y_{i,j}^r\right) $ is defined as a point in $ E \setminus \{ p_i^e \} \cup C $. Additionally, $ x_{i,j}^r $ falls within the interval $ \left[x_i^e - \tau, x_i^e + \tau\right] $ and $ y_{i,j}^r $ within $ \left[y_i^e - \tau, y_i^e + \tau\right] $, where $ \tau \in \mathbb{Z}^+$.
In particular, MPE only considers endpoints as potential $ P_h $, excluding junction points where multiple strokes converge, as manipulating these would compromise character structural integrity. Each endpoint $ p_i^e $ is associated with multiple reference points $ p_{i,j}^r$, collectively denoted as $ R_i = \left\{ p_{i,j}^r \middle| \ 0 \leq j \leq m \right\} $, where $ m $ represents the cardinality of $ E \setminus \{ p_i^e \} \cup C $.

\noindent\textit{Overall Handle-point Selection.}
After establishing $ R_i $ for each endpoint $ p_i^e $, MPE assigns a score $ s_{i,j} $ to each $ p_{i,j}^r $ in $ R_i $, based on specific rules below, which are recorded as triplets $ \left( p_i^e, p_{i,j}^r, s_{i,j} \right) $. A higher score $ s_{i,j} $ indicates a greater suitability of $ p_i^e $ as the handle point. The $ p_i^e $ associated with the highest score $ s_{i,j} $ is designated as $ P_h $, with its corresponding $ p_{i,j}^r $ established as the reference point $ P_r $ for $ P_h $.

\noindent\textit{Scoring and Tie-breaking Rules.}
We hereby describe our scoring rules. For an endpoint $ p_i^e $ with $ |R_i| = 0$ (no reference point), the score is set to 0. When $ |R_i| > 0$, each $ p_{i,j}^r $ receives an initial score of 1. MPE evaluates each pair $ \left(p_i^e, p_{i,j}^r\right)$ by analyzing their connectivity, determining whether they lie on the same stroke. If $p_i^e$ and $ p_{i,j}^r $ share the same stroke (denoted as $ p_i^e \sim p_{i,j}^r $), the score remains unchanged; otherwise, it increases by one. When multiple $ p_{i,j}^r $s achieve a score of 2, an additional point is awarded to $ p_{i,j}^r $ with the smallest y-coordinate. This procedure is formalized as:
\begin{equation}
	\mathcal{R}^1 =
	\begin{cases}
		1, & \text{if $ p_i^e \nsim p_{i,j}^r $}\\
		0, & \text{if $ p_i^e \sim p_{i,j}^r $}
	\end{cases}, \quad s_{i,j} = 1+\mathcal{R}^1.
\end{equation}
\begin{equation}
	\mathcal{R}^2\! =\!
	\begin{cases}
		1,\! & \!\text{if } y^r_{i,j}\! = \!\min \left\{ y^r_{i,p}| s_{i,p} = 2, \ 0\!< \! \! p \!<m \!-\! 1\right\} \\
		0, \!& \text{otherwise}
	\end{cases},
\end{equation}

Let $ \widehat{p}^r_{i} $ denote the point with the smallest y-coordinate in $ \mathcal{R}^2 $, with its corresponding score $ \widehat{s}_{i} $. The application of the aforementioned rules generates a set of tuples, recorded as $ (p^e_{i}, \widehat{p}^r_{i}, \widehat{s}_{i}) $. The endpoint $ p^e_i $ associated with the highest score is designated as $ P_h $. In cases where multiple endpoints share the maximum score, the endpoint with the smallest y-coordinate is selected as the final $ P_h $. Overall, MPE outputs a handle/reference-point pair in which the handle point is a safely movable endpoint and the pair can be consistently recovered during extraction.

\vspace{0.3em}
\noindent \textbf{Target Point Estimation.}
\name{} embeds information by moving the handle point $ P_h $ to the corresponding point $P_t $. In the previous part, we thoroughly describe the selection process of $ P_h $. We hereby elaborate on the specific methodology for determining the corresponding $ P_t $.

\noindent\textit{Encoding Axis and Movement Conditions.}
Assuming the relative distance between $ P_h $ and $ P_r $ along the $ \lambda $-axis is denoted as $ \Delta\left(P_h, P_r\right) $, where $ \lambda $-axis can be modeled as follows: 
\begin{equation}
	\lambda\text{-axis} \text{ is} \begin{cases}
		\text{X-axis}, &d_x \leq d_y\\
		\text{Y-axis}, &d_x > d_y\\
	\end{cases},
\end{equation}
Here, $ d_x $ and $ d_y $ can be calculated via
\begin{equation}
	\label{dxdy}
	d_x = \left| x_h - x_r\right|, \ d_y = \left| y_h - y_r\right|,
\end{equation}
where $ \left(x_h, y_h\right) $, $ \left(x_r, y_r\right) $ are the coordinates of $ P_h $ and $ P_r $, respectively, and $ \Delta\left(P_h, P_r\right) = \min\{d_x, d_y\}$. To embed a bit value of 0, if $ \Delta\left(P_h, P_r\right) > T_{\text{embed}} $, $ P_h $ is relocated to ensure $ \Delta\left(P_t, P_r\right) \leq T_{\text{embed}}$. If $ \Delta\left(P_h, P_r\right) \leq T_{\text{embed}}$ already, no repositioning is needed, thus avoiding unnecessary distortion. To embed a bit value of 1, $ P_h $ is relocated if $ \Delta\left(P_h, P_r\right) \leq T_{\text{embed}}$ to ensure $ \Delta\left(P_t, P_r\right) > T_{\text{embed}}$; otherwise, no adjustment is required.

\noindent\textit{Handle-point Identity and Synchronization.}
Since $ P_t $ is relocated from $ P_h $, we need to ensure that $ P_h $ maintains its identity from the embedding stage to the extraction stage when determining the position of $ P_t $. Otherwise, synchronization errors may occur, compromising message recovery, \eg, if the y-coordinate of $ P_h $ increases during embedding and exceeds that of other endpoints $p_i^e$ within the triplet $ (p_e^{i}, \widehat{p}_r^{i}, \widehat{s}^{i}) $ sharing the same score, the decoder may mistakenly treat an unmoved $p_i^e$ as $ P_h $. 

\noindent\textit{Target-point Calculation and Special Cases.}
To address this problem, we analyze the computation of $ P_t $ based on the embedded bit value (0 or 1). When the hidden bit is 0, $ P_t $'s position depends on both the stroke direction containing $ P_h $ and the direction from $ P_h $ to $ P_r $, represented by $\vec{\mathcal{V}}$ and $\vec{\mathcal{H}}$, respectively. To calculate $\vec{\mathcal{V}}$, we first construct a circle centered at $ P_h $ with radius $ r_h $, as illustrated by the blue dashed circle in Fig.~\ref{fig:pipeline}. Designating the intersection of this circle with $I_{\text{ske}}$ as $ P_k $ with coordinates $ \left(x_k, y_k\right) $, $\vec{\mathcal{V}}$ can be computed as 
\begin{equation}
	\label{v}
	\vec{\mathcal{V}} = \left(\mathcal{V}_x, \mathcal{V}_y\right) = \left(x_h-x_k, \ y_h-y_k\right).
\end{equation}

Similarly, $\vec{\mathcal{H}}$ is calculated as Eq.~\ref{v}.
Then, $P_t \left(x_t, y_t\right)$ moved a distance $ \mathcal{D} $ can be expressed as:
\begin{equation}
\label{pt}
x_t = x_h + \mathcal{D} \times \frac{\mathcal{V}_x}{\|\vec{\mathcal{V}}\|} \times \text{sgn}\left(\mathcal{H}_x\right), 
y_t = y_h + \mathcal{D} \times \frac{\mathcal{V}_y}{\|\vec{\mathcal{V}}\|} \times \text{sgn}\left(\mathcal{H}_y\right),
\end{equation}
where $ \mathcal{D} $ is defined in Eq.~(\ref{d}), and $\text{sgn}\left(\cdot\right) $ is a sign function.
\begin{equation}
	\label{d}
	\mathcal{D} = \begin{cases}
		|x_h-x_r|\ / \ |\mathcal{V}_x|, &\text{if } \lambda-\text{axis is X-axis}\\
		|y_h-y_r|\ / \ |\mathcal{V}_y|, &\text{if } \lambda-\text{axis is Y-axis}
	\end{cases}.
\end{equation}

When the embedding bit is 1, it is necessary to increase $ \Delta\left(P_h, P_r\right) $ along $ \lambda$-axis to complete the embedding process. The calculation for $ \left(x_t,y_t\right) $ is similar to the aforementioned steps. We simply need to multiply $ \text{sgn}(x) $ in Eq.~(\ref{pt}) by -1 to indicate that $ P_h $ is moving away from $ P_r $. 

Beyond the above cases, additional scenarios warrant further analysis. 
When the $ \lambda $-axis corresponds to the $ x $-axis and $ \Delta\left(P_h, P_r\right) =0$, we designate that $P_h$ moves along the direction of stroke elongation.
When $ \lambda$-axis aligns with the $ y $-axis and $ \Delta\left(P_h, P_r\right)=0$, 
we calculate the angle $ \theta $ between $ \vec{\mathcal{V}} $ and the positive x-axis using Eq.~(\ref{theta}). 
\begin{equation}
\label{theta}
	\theta = \arccos\frac{\mathcal{V}_x}{\|\vec{\mathcal{V}}\|}, \ \theta \in [0, 2\pi ).
\end{equation}

\begin{wrapfigure}{r}{0.51\textwidth}
\centering
\vspace{-1.5em}
\includegraphics[width=0.49\textwidth]{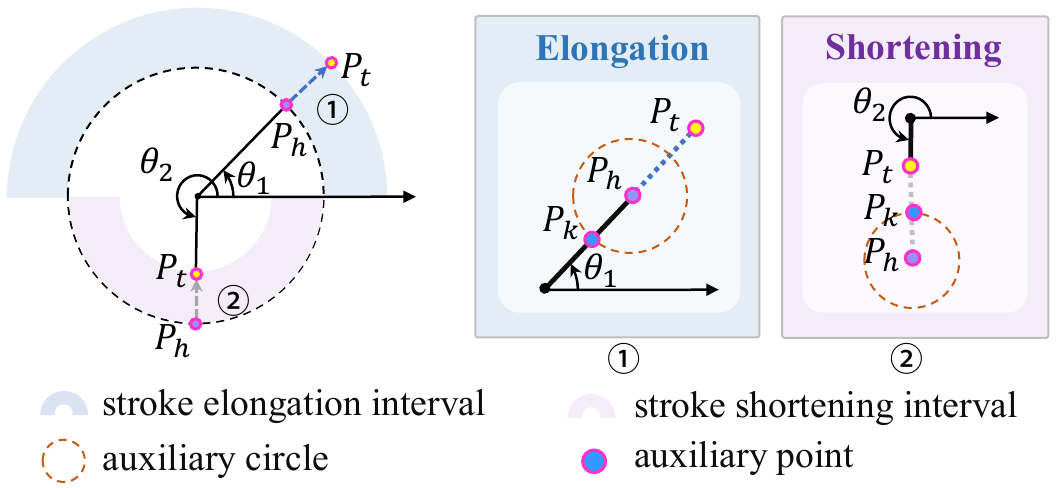}
\caption{Illustration the direction of $ P_h$ movements corresponding to its stroke in various orientations. }
\vspace{-1em}
\label{fig:stroke}
\end{wrapfigure}
When $ 0 \le \theta < \pi $, we increase $ \Delta\left(P_h, P_r\right) $ by elongating the stroke; otherwise, we shorten it. Fig.~\ref{fig:stroke} illustrates the moving direction of $P_h $ corresponding to different intervals of $\theta$. As shown in the blue box of Fig.~\ref{fig:stroke}, we first calculate $ \vec{\mathcal{V}} $ using an auxiliary circle, then derive angle $ \theta_1 $ from $ \vec{\mathcal{V}} $. Since $ \theta_1 < 180^\circ$, we move $ P_h $ along $ \vec{\mathcal{V}} $ towards the upper right. We can observe that the y-coordinate of $ P_t $ is smaller than the y-coordinate of $ P_h $, which ensures the identity of $P_h $. Overall, TPE outputs a target point that satisfies the intended bit condition with minimal, stroke-aligned movement while preserving character identity and extraction consistency.

\vspace{0.3em}
\noindent \textbf{Mask Drawing Module.}
After obtaining the handle points and target points, we can construct the mask $ \mathcal{M} $. In general, it is a binary image of size $ H \times W $, where the editing region is represented by a `rectangle' (with a pixel value 255), \ie, $ \Omega=\left\{\left(x,y\right): \Gamma_{\text{left}}\! -\! \sigma \leq x \leq \Gamma_{\text{right}}\! +\! \sigma, \Gamma_{\text{top}}\! -\! \sigma \leq y\leq\Gamma_{\text{bottom}}\! +\! \sigma\right\} $, where $ \sigma \in \mathbb{Z}^+ $. Specifically, taking the case where the $ \lambda\text{-axis} $ corresponds to the $ x\text{-axis} $ as an example, we first compute $ \Gamma_{\text{left}} $ and $ \Gamma_{\text{right}} $ via:
\begin{equation}
	\Gamma_{\text{left}} = \min(x_h, x_t),
	\ \Gamma_{\text{right}} = \max(x_h, x_t).
\end{equation}
Then, we extract a candidate area $ \mathcal{M}_c $ from $ I_{cover} $, defined by a four-tuple $ (0, \Gamma_{\text{left}}, H, \Gamma_{\text{right}}-\Gamma_{\text{left}}) $ that specifies its top-left corner $ (0, \Gamma_{\text{left}}) $ and its height and width $ (H, \Gamma_{\text{right}}-\Gamma_{\text{left}}) $, to calculate $\Gamma_{\text{top}} $ and $ \Gamma_{\text{bottom}} $. The detailed calculation process is shown in Appendix~\ref{sec:appdixMDM}.  
Thus, we have obtained $ \Gamma_{\text{top}} $, $\Gamma_{\text{bottom}}$, $\Gamma_{\text{left}}$, and $ \Gamma_{\text{right}} $. 
Although we aim to minimize the mask size to reduce interference with the original image, excessively small masks may compromise the diffusion process, potentially producing outputs identical to the original image. Consequently, we expand $ \mathcal{M} $ by extending each boundary by a distance of $ \sigma $, enlarging the mask area and improving diffusion quality.

\subsection{Guided Diffusion Encoding}
\label{diffusion}
After obtaining the handle point $ P_h $, the target point $ P_t $, and the mask $ \mathcal{M} $, we exploit DragDiffusion to move $ P_h $ to $ P_t $. The process involves three key phases. Initially, DragDiffusion fine-tunes the diffusion model's UNet with LoRA to better capture the features of the original image, enhancing identity consistency throughout the editing process.  Following this, we apply DDIM inversion to $ I_{\text{cover}} $ to generate an initial diffusion latent that serves as the foundation for latent optimization. This optimization, guided by $ P_h $ and $ P_t $, iteratively applies motion supervision and point tracking until the handle point aligns with its target. Finally, DragDiffusion implements reference-latent control during denoising by replacing the key and value vectors of the self-attention in the edited latent with those from the initial latent, thereby directing attention toward consistent features and textures.

DragDiffusion controls the update of the latent $ z_t $ (the result of the $t$-th step of DDIM inversion) via the motion supervision loss $ L_{ms}$. However, our experiments (in Section~\ref{sec: ablation}) reveal that content generated within the mask tends to cause significant changes in local shapes. As such, to satisfy invisibility requirements, we introduce a local consistency loss $ L_{lc} $. Denoting the feature vector at pixel location $ P_h $ as $ G_{P_h}(\cdot) $, the loss at the $ k $-th iteration is given by:
\begin{equation}
	L_{lc}(\hat{z}_t^k) = \sum_{q \in \Omega} \left\| G_{q+d}(\hat{z}_{t-1}^k) - \text{sg}(G_q(\hat{z}_{t-1}^0)) \right\|_1, \qquad
	\label{lossall}
	L(\hat{z}_t^k) = L_{ms}(\hat{z}_t^k) + \eta L_{lc}(\hat{z}_t^k),
\end{equation}
where $ \hat{z}_t^k $ is the latent variable at step $ t $ after the $ k $-th update, $ \text{sg}(\cdot) $ is the stop-gradient operator, vector $ d = h^k-h^0 $ denotes the displacement vector, where $h^k$ is the handle point at the $k$-th motion supervision iteration, $\Omega$ is the editing region defined by MDM, and $\eta$ is weighting coefficient. 

\subsection{Masked Region Replacement}
\label{sec-mrr}
In our experiments (Section~\ref{sec: ablation}), we observe that DragDiffusion may cause feature alterations beyond the target region. To preserve the original features outside the edited region, we propose a masked region replacement method that replaces the content within the masked region of $I_{cover} $ with corresponding content from the generated image. With this single step, information embedding is achieved while minimizing the impact on the original character.

\subsection{Document Embedding and Extraction}
To improve embedding efficiency, \name{} maintains a
glyph cache (referred to as a codebook) that stores
mappings between previously processed glyphs and their
automatically generated watermarked variants for bits
0 and 1. When a previously processed glyph is encountered,
its corresponding variant is directly retrieved from
the cache. For unseen glyphs or styles, \name{} automatically
generates the corresponding variants and adds them to
the cache for future reuse, without manual intervention
or additional font-specific training. Unlike traditional
font-based codebooks, this cache serves solely as an
efficiency mechanism rather than a prerequisite for
watermark embedding. Notably, watermark extraction
does not rely on the codebook.

Once we obtain the document with the embedded information, we implement the following extraction procedure. First, we segment individual characters using the CRAFT algorithm \cite{baek2019character}, following the approach adopted in AutoStegaFont. Then, we apply the Movement Probability Estimator (MPE) to the structural representation of each character to identify $P_h$ and $P_r$ and calculate the distance $ \Delta\left(P_h, P_r\right)^{'} $ between these two points. The extraction decision follows a simple threshold rule: if $ \Delta\left(P_h, P_r\right)^{'} > T_{\text{embed}} $, we extract a bit value of `1'; otherwise, a bit of `0' is extracted.

\section{Experiment}
\label{sec:Experiment}

\subsection{Main Settings}
\label{subsec:settings}

\noindent \textbf{Datasets.} 
We conduct our main experiments on five widely used English and Chinese fonts. For English, we select Calibri, Arial, and Times New Roman (TNR). For Chinese, we use SimSun and SimHei. These fonts cover diverse typographic styles and stroke structures. Each font is evaluated at six sizes (in pt): {12, 16, 20, 24, 28, 36}.
To further assess generalizability, we evaluate \name{} on handwritten and artistic fonts, as well as other languages and mathematical formulas. Detailed results for the latter two settings are provided in Appendix~\ref{subsec:appdixforeign}--\ref{subsec:formula}.

\vspace{0.3em}
\noindent \textbf{Baseline Selection.}
Since \name{} modifies the visual appearance of characters
without altering the textual content or document layout,
font-based and image-based methods are the most relevant
baselines. We select AutoStegaFont (ASF) as a representative
font-based method, and StegaStamp and IHA as representative
image-based methods. In main experiments, we compare \name{} with the state-of-the-art (SOTA) font-based document hiding method AutoStegaFont (ASF) \cite{yang2023autostegafont} on the datasets described above. We also compare \name{} with two image-based methods, StegaStamp \cite{tancik2020stegastamp} and IHA \cite{yang2023language}, on document images, with visual analysis presented in Appendix~\ref{compareAppendix}. To ensure a fair evaluation, we maintain consistent experimental equipment (\eg, printers, scanners) and environmental conditions across all methods. More implementation details are in Appendix~\ref{implementationAppendix}.

\vspace{0.3em}
\noindent \textbf{Settings for Document Hiding.}
In our method, the parameter $ \eta $ in Eq.~(\ref{lossall}) is set to 0.003, and $ T_{\text{embed}} $ is set to 10. When the shortest edge of the original image is 512, $ \tau $ and $ \sigma $ are set to 80 and 10, respectively. Their values are adjusted based on the image size. We adopt DragDiffusion with the same parameters as specified in \cite{shi2024dragdiffusion}. As for ASF, we follow the same settings described in \cite{yang2023autostegafont} and train on multiple fonts to generate character codebooks.

\vspace{0.3em}
\noindent \textbf{Evaluation Metrics.}
We evaluate robustness using average extraction accuracy (ACC), which measures the percentage of correctly extracted bits, with higher values indicating stronger robustness. To assess stealthiness, we adopt two widely used image quality metrics: Peak Signal-to-Noise Ratio (PSNR) and Structural Similarity Index (SSIM) \cite{wang2004image}, with higher values corresponding to better visual imperceptibility. Efficiency-related discussions are detailed in Appendix~\ref{sec:appdixefficiency}.

\subsection{Main Results}
\label{subsec:mainResult}

\begin{wrapfigure}{r}{0.55\textwidth}
	\centering
	\vspace{-0.5em}
	\includegraphics[width=0.5\textwidth]{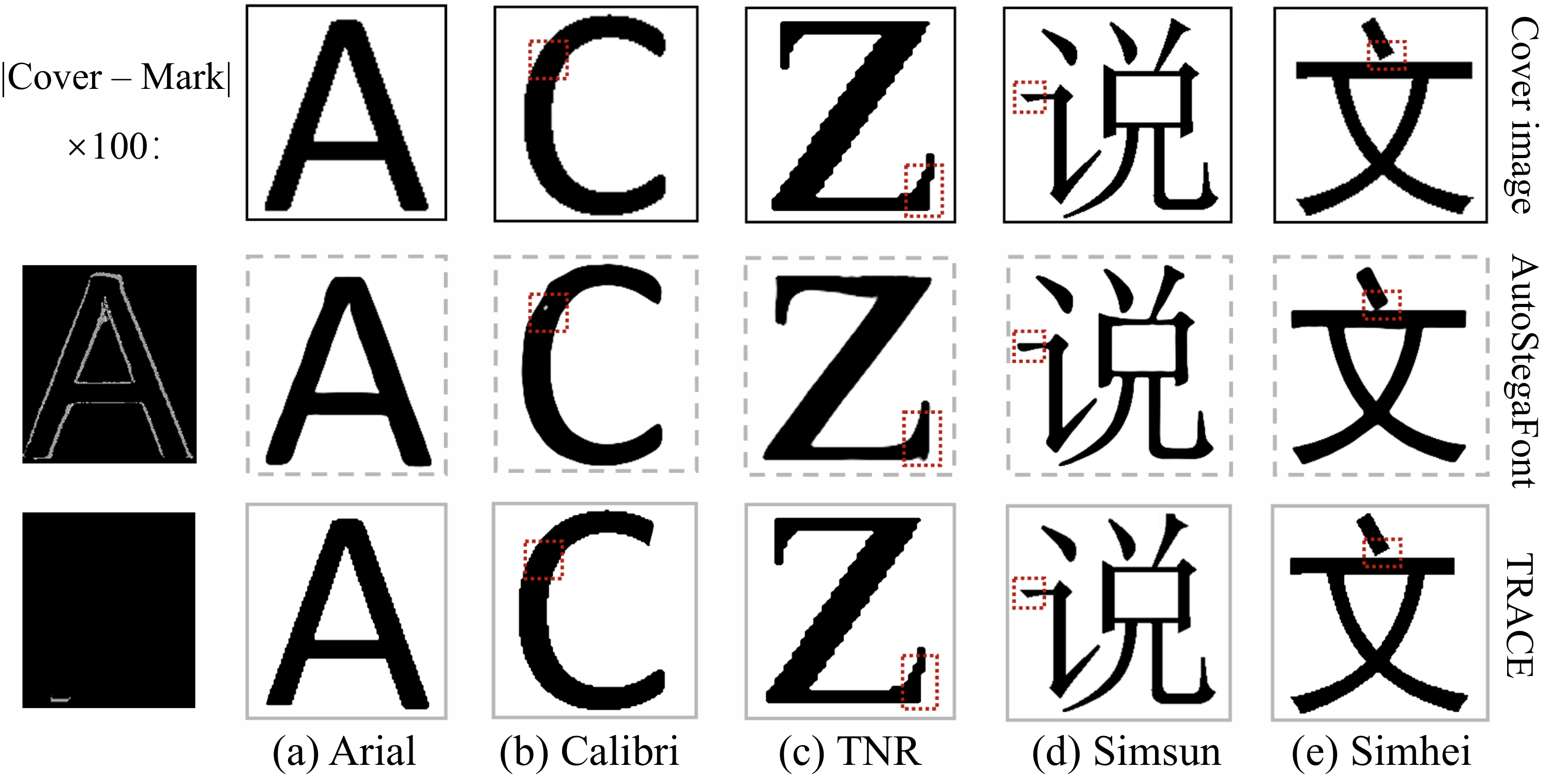}
	\caption{Visual comparisons of encoded images of our \name{} and the comparison method \cite{yang2023autostegafont}.}
	\label{fig:visualcompare}
	\vspace{-1em}
\end{wrapfigure}

\vspace{0.3em}
\noindent \textbf{Performance of Imperceptibility.}
Fig.~\ref{fig:visualcompare} presents a comparison of encoded images generated by 
our \name{}, and ASF. The encoded image produced by \name{} demonstrates virtually imperceptible differences from the original image, with the residual image appearing almost completely black. 
In contrast, ASF's encoded images display noticeable alterations along the edges, clearly visible in the residual image. The letter ``C" in particular exhibits a hollowing effect in the ASF encoding. Quantitatively, \name{} achieves superior PSNR and SSIM metrics across both Chinese and English fonts, as detailed in Tab.~\ref{tab:psnrssim}.

\begin{table}[h]
	\centering
	\caption{PSNR and SSIM of \name{} and ASF, with the best results marked in \textbf{boldface}.}
	\label{tab:psnrssim}
	\setlength\tabcolsep{3.5pt}{
    \small
	\begin{tabular}{c@{\hspace{10pt}}c@{\hspace{10pt}}ccccc}
	\toprule
	\multirow{2}{*}{} & Method & Arial & TNR & Calibri & Simsun & Simhei\\
	\midrule
	\multirow{2}{*}{PSNR$\uparrow$} 
	& ASF & 26.56 & 27.29 & 27.46 & 30.34 & 26.23 \\
	& \name{} & \textbf{36.76} & \textbf{36.51} & \textbf{32.38} & \textbf{40.98} & \textbf{40.05} \\
	\midrule
	\multirow{2}{*}{SSIM$\uparrow$} 
	& ASF & 0.984 & 0.986 & 0.985 & 0.989 & 0.986 \\
	& \name{} & \textbf{0.996} & \textbf{0.995} & \textbf{0.995} & \textbf{0.998} & \textbf{0.998} \\
	\bottomrule
	\end{tabular}
	}
	\vspace{-0.5em}
\end{table}

\begin{table*}[!t]
	\centering
	
	\begin{minipage}[t]{0.48\textwidth}
	\centering
	\caption{Robustness against Screenshots across varying character sizes.}
	\vspace{-0.5em}
	\label{tab:screenshots}
    \renewcommand{\arraystretch}{1.2}
	\scalebox{0.67}{
	\begin{tabular}{c@{\hspace{5pt}}c@{\hspace{10pt}}cccccc}
	\toprule
	\multirow{2}{*}{\raisebox{-0.6ex}[0pt][0pt]{Font}} & \multirow{2}{*}{\raisebox{-0.6ex}[0pt][0pt]{Method}} & \multicolumn{6}{c}{\textbf{Screenshots}} \\
	\cmidrule(lr){3-8}
	& & 12pt & 16pt & 20pt & 24pt & 28pt & 36pt \\
	\midrule
	\multirow{2}{*}{\raisebox{-0.6ex}[0pt][0pt]{\textbf{Arial}}} 
	& ASF & 85.83 & 91.67 & 90.00 & 87.50 & 88.33 & 82.50 \\
	& \name{} & \textbf{96.67} & \textbf{97.50} & \textbf{99.17} & \textbf{100} & \textbf{100} & \textbf{100} \\
	\hline
	\multirow{2}{*}{\raisebox{-0.6ex}[0pt][0pt]{\textbf{TNR}}} 
	& ASF & 70.83 & 77.50 & 72.50 & 69.17 & 74.17 & 80.83 \\
	& \name{} &\textbf{92.50} & \textbf{92.50} & \textbf{95.00} & \textbf{97.50} & \textbf{100} & \textbf{100} \\
	\hline
	\multirow{2}{*}{\raisebox{-0.6ex}[0pt][0pt]{\textbf{Calibri}}} 
	& ASF & 95.00 & 95.83 & 96.67 & 98.33 & 97.50 & \textbf{100} \\
	& \name{} & \textbf{97.50} & \textbf{99.17} & \textbf{99.17} & \textbf{100} & \textbf{100} & \textbf{100} \\
	\hline
	\multirow{2}{*}{\raisebox{-0.6ex}[0pt][0pt]{\textbf{Simsun}}} 
	& ASF & 86.67 & 87.50 & 90.83 & 91.67 & 92.50 & 93.33 \\
	& \name{} & \textbf{90.83} & \textbf{91.67} & \textbf{95.00} & \textbf{95.83} & \textbf{95.83} & \textbf{97.50} \\
	\hline
	\multirow{2}{*}{\raisebox{-0.6ex}[0pt][0pt]{\textbf{Simhei}}} 
	& ASF & 91.67 & 92.50 & 94.17 & 95.83 & 97.50 & 98.33 \\
	& \name{} & \textbf{92.50} &\textbf{95.83} & \textbf{96.67} & \textbf{97.50} & \textbf{97.50} & \textbf{99.17} \\
	\bottomrule
	\end{tabular}
	}
	\end{minipage}
	\hfill
	\begin{minipage}[t]{0.48\textwidth}
	\centering
	\caption{Robustness against Print-scan across varying character sizes.}
	\vspace{-0.5em}
	\label{tab:print_scan}
    \renewcommand{\arraystretch}{1.2}
	\scalebox{0.67}{
	\begin{tabular}{c@{\hspace{5pt}}c@{\hspace{10pt}}cccccc}
	\toprule
	\multirow{2}{*}{\raisebox{-0.6ex}[0pt][0pt]{Font}} & \multirow{2}{*}{\raisebox{-0.6ex}[0pt][0pt]{Method}} & \multicolumn{6}{c}{\textbf{Print Scans}} \\
	\cmidrule(lr){3-8}
	& & 12pt & 16pt & 20pt & 24pt & 28pt & 36pt \\
	\midrule
	\multirow{2}{*}{\raisebox{-0.6ex}[0pt][0pt]{\textbf{Arial}}} 
	& ASF & 80.83 & 68.33 & 70.83 & 72.50 & 70.00 & 76.67 \\
	& \name{} & \textbf{95.83} & \textbf{97.50} & \textbf{99.17} & \textbf{99.17} & \textbf{100} & \textbf{100} \\
	\hline
	\multirow{2}{*}{\raisebox{-0.6ex}[0pt][0pt]{\textbf{TNR}}} 
	& ASF & 64.17 & 72.50 & 79.17 & 68.33 & 79.17 & 72.50 \\
	& \name{} & \textbf{92.50} &\textbf{94.17} & \textbf{95.83} & \textbf{97.50} & \textbf{99.17} & \textbf{99.17} \\
	\hline
	\multirow{2}{*}{\raisebox{-0.6ex}[0pt][0pt]{\textbf{Calibri}}} 
	& ASF & 73.33 & 77.50 & 75.00 & 70.00 & 69.17 & 74.17 \\
	& \name{} & \textbf{98.33} & \textbf{99.17} & \textbf{100} & \textbf{100} & \textbf{100} & \textbf{100} \\
	\hline
	\multirow{2}{*}{\raisebox{-0.6ex}[0pt][0pt]{\textbf{Simsun}}} 
	& ASF & 75.83 & 75.00 & 74.17 & 75.83 & 76.70 & 78.33 \\
	& \name{} & \textbf{89.17} & \textbf{91.67} & \textbf{94.17} & \textbf{95.00} & \textbf{96.67} & \textbf{98.33} \\
	\hline
	\multirow{2}{*}{\raisebox{-0.6ex}[0pt][0pt]{\textbf{Simhei}}} 
	& ASF & 77.50 & 87.50 & 84.17 & 85.00 & 88.33 & 87.50 \\
	& \name{} & \textbf{91.67} & \textbf{92.50} & \textbf{95.00} & \textbf{96.67} & \textbf{97.50} & \textbf{98.33} \\
	\bottomrule
	\end{tabular}
	}
	
	\end{minipage}
	
	\vspace{1em}
	
	\begin{minipage}[t]{0.48\textwidth}
	\centering
	\caption{Robustness against Print-camera captures across varying sizes.}
	\vspace{-0.5em}
	\label{tab:print_camera}
    \renewcommand{\arraystretch}{1.2}
	\scalebox{0.67}{
	\begin{tabular}{c@{\hspace{5pt}}c@{\hspace{10pt}}cccccc}
	\toprule
	\multirow{2}{*}{\raisebox{-0.6ex}[0pt][0pt]{Font}} & \multirow{2}{*}{\raisebox{-0.6ex}[0pt][0pt]{Method}} & \multicolumn{6}{c}{\textbf{Print Camera}} \\
	\cmidrule(lr){3-8}
	& & 12pt & 16pt & 20pt & 24pt & 28pt & 36pt \\
	\midrule
	\multirow{2}{*}{\raisebox{-0.6ex}[0pt][0pt]{\textbf{Arial}}} 
	& ASF & 66.67 & 70.83 & 76.67 & 77.50 & 72.50 & 80.00 \\
	& \name{} & \textbf{95.83} & \textbf{96.67} & \textbf{98.33} & \textbf{98.33} & \textbf{99.17} & \textbf{100} \\
	\hline
	\multirow{2}{*}{\raisebox{-0.6ex}[0pt][0pt]{\textbf{TNR}}} 
	& ASF & 65.83 & 75.00 & 71.67 & 70.83 & 77.50 & 75.83 \\
	& \name{} & \textbf{90.00} & \textbf{91.67} & \textbf{94.17} & \textbf{94.50} & \textbf{98.33} & \textbf{98.33} \\
	\hline
	\multirow{2}{*}{\raisebox{-0.6ex}[0pt][0pt]{\textbf{Calibri}}} 
	& ASF & 71.67 & 72.50 & 75.83 & 69.17 & 71.67 & 72.50 \\
	& \name{} & \textbf{96.67} & \textbf{97.50} & \textbf{97.50} & \textbf{99.17} & \textbf{100} & \textbf{100} \\
	\hline
	\multirow{2}{*}{\raisebox{-0.6ex}[0pt][0pt]{\textbf{Simsun}}} 
	& ASF & 70.83 & 75.83 & 73.33 & 75.00 & 74.17 & 76.67 \\
	& \name{} & \textbf{85.83} & \textbf{86.67} & \textbf{89.17} & \textbf{91.67} &\textbf{94.17} & \textbf{96.67} \\
	\hline
	\multirow{2}{*}{\raisebox{-0.6ex}[0pt][0pt]{\textbf{Simhei}}} 
	& ASF & 77.50 & 72.50 & 76.67 & 79.17 & 83.33 & 85.83 \\
	& \name{} & \textbf{90.00}& \textbf{90.83} & \textbf{93.33} & \textbf{95.00} & \textbf{95.83} & \textbf{96.67} \\
	\bottomrule
	\end{tabular}
	}
	\end{minipage}
	\hfill
	\begin{minipage}[t]{0.48\textwidth}
	\centering
	\caption{Invisibility and Robustness Comparison of \name{} and baselines on document images, evaluating robustness to screenshots, print-scan, and print-camera.}
	\vspace{-0.5em}
	\label{tab:compare2}
    \renewcommand{\arraystretch}{1.2}
	\small
	\scalebox{0.96}{
	\begin{tabular}{lccc}
	\toprule
	Method & Stegastamp & IHA & \name{}\\
	\midrule
	Screenshot & \textbf{100} & 84.58 & \textbf{100} \\
	Print-scan & 98.54 & 84.29 & \textbf{99.05} \\
	Print-camera & 98.12 & 83.94 & \textbf{98.75} \\
	PSNR & 27.19 & 29.60 & \textbf{33.34} \\
	SSIM & 0.8986 & 0.9910 & \textbf{0.9962} \\
	\bottomrule
	\end{tabular}
	}
	\end{minipage}
\end{table*}
\vspace{0.3em}
\noindent \textbf{Performance of Robustness.}
We comprehensively evaluate \name{}'s robustness against various real-world distortions compared to ASF \cite{yang2023autostegafont}, with results recorded in Tab.~\ref{tab:screenshots}-~\ref{tab:print_camera}. \name{} consistently outperforms ASF across all testing conditions, even achieving error-free extraction in some cases. This performance advantage is particularly pronounced for smaller font sizes and challenging print-scan and print-camera scenarios. Fig.~\ref{fig:visualcompare} reveals that ASF's limited robustness stems from its reliance on edge pixel manipulation, which is highly vulnerable to distortions. For instance, screenshots blur text edges, while printing introduces substantial noise along edges. In contrast, \name{} embeds watermarks through subtle structural adjustments that preserve the core character geometry despite edge pixel degradation. Note that both \name{} and ASF can correctly extract information without attacks. 
We compare our method with Stegastamp \cite{tancik2020stegastamp} and IHA \cite{yang2023language} on Calibri documents, both of which support cross-media transmission of text images. Table~\ref{tab:compare2} reports quantitative results on robustness and invisibility. \name{} consistently achieves the best performance across all metrics, demonstrating superior robustness and invisibility. Visual comparisons are provided in Appendix~\ref{compareAppendix}. 


\begin{wrapfigure}{r}{0.45\textwidth}
\centering
\vspace{-1em}
\includegraphics[width=0.443\textwidth]{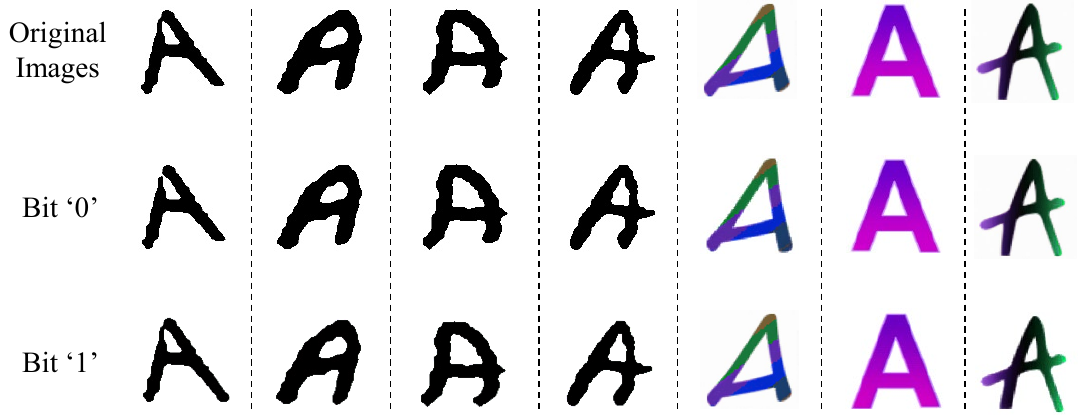}
\vspace{-0.5em}
\caption{Generalizability of our \name{} from handwritten to artistic fonts.}
\label{fig:handart}
\vspace{-1em}
\end{wrapfigure}
\begin{figure*}[h]
    \centering
    \begin{minipage}[c]{0.37\textwidth} 
        \centering
       \includegraphics[width=\textwidth]{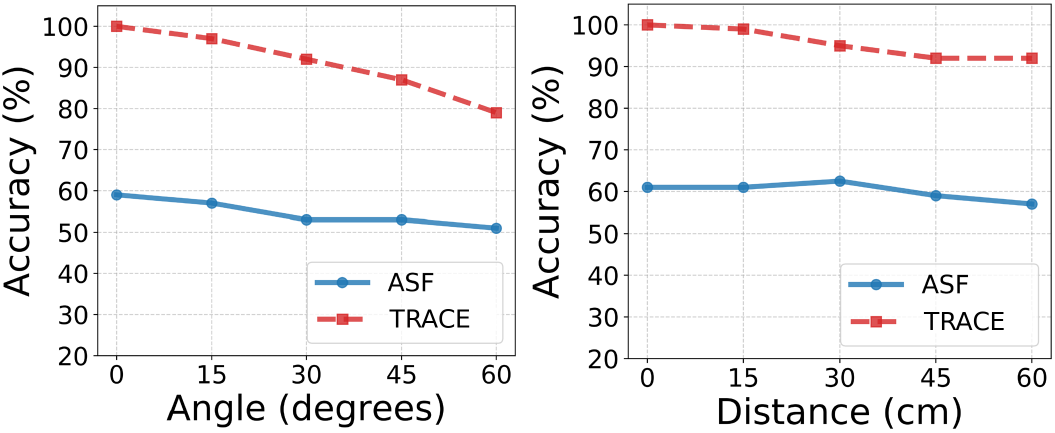}
        \caption{Robustness against shooting with different angles and distances, tested with Calibri 12pt.}
        \label{rotate}
    \end{minipage}
    \hspace{0.2em}
    \begin{minipage}[c]{0.6\textwidth} 
    \vspace{-0.9em}
        \centering
        \includegraphics[width=\textwidth]{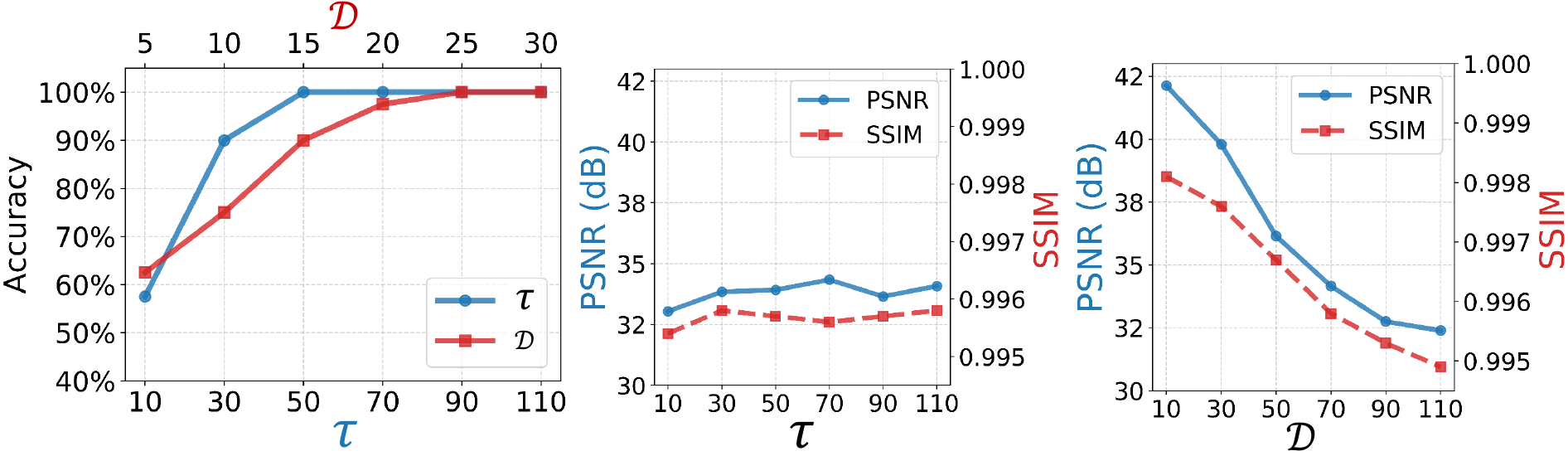}
        \caption{Hyper-parameter analysis for $\mathcal{D}$ and $\tau$ for robustness (ACC), and invisibility (PSNR, SSIM) using 100 Calibri characters of varying sizes.}
        \label{hyperparameter}
    \end{minipage}
    \vspace{-1.5em}
\end{figure*}

Given that photographing paper documents is common practice, we evaluated robustness under varying shooting distances and angles, as shown in Fig.~\ref{rotate}. The left graph shows that \name{} maintains high accuracy ($>80\%$) even as the shooting angle increases to $60^\circ$, while ASF's performance declines to approximately 50\%. The right graph shows that \name{} sustains consistently high accuracy ($>90\%$) across different shooting distances (20-60 cm), whereas ASF exhibits lower performance (approximately 60\%) that slightly decreases with distance.

\vspace{0.3em}
\noindent \textbf{Performance of Generalizability.}
We embed both 0-bit and 1-bit data into diverse handwritten and artistic fonts. As shown in Fig.~\ref{fig:handart}, \name{} demonstrates strong generalizability across various font styles. We further assess the robustness and invisibility of \name{} on these font types, with results presented in Table~\ref{tab:handart}. We can see that \name{} achieves superior performance in both robustness and invisibility, striking a favorable balance among generalizability, robustness, and invisibility, which represents a significant advancement over previous methods.
\begin{table}[h]
	\centering
	\caption{Invisibility and robustness of \name{} on handwritten and artistic fonts under screenshots, print-scan, and print-camera distortions.}
	\label{tab:handart}
	\small
	\setlength{\tabcolsep}{3.5pt}
	\begin{tabular}{cccccc}
	\toprule
	Method & Screenshots & Print-scan & Print-camera & PSNR & SSIM\\
	\midrule
	Handwritten & 94.43 & 93.17 & 91.67 & 38.20 & 0.996 \\
	Artistic    & 97.27 & 94.77 & 92.93 & 41.37 & 0.995 \\
	\bottomrule
	\end{tabular}
\end{table}

\subsection{Ablation Study}
\label{sec: ablation}
\noindent \textbf{Effects of Parameters $\mathcal{D}$ and $\tau$.}
The hyper-parameter $\mathcal{D}$ controls the degree of handle point movement. If $\mathcal{D}$ is too small, it may inadequately guide DragDiffusion in generating encoded images. Conversely, excessive $\mathcal{D}$ values can produce overly noticeable movements, compromising invisibility. As shown in Fig.~\ref{hyperparameter}, except for these extreme cases, \name{} achieves an optimal balance between robustness and invisibility, offering a broad effective implementation range. Regarding hyper-parameter $\tau$, a too small $\tau$ may cause a mismatch between $P_r$ values of $P_h$ before and after movement. However, as shown in Fig.~\ref{hyperparameter}, our method's invisibility is not sensitive to variations in $\tau$. As such, a wide range of $\tau$ can be selected for \name{}.

Additional ablation studies on MPE, TPE, $L_{lc}$, and MRR are provided in Appendix~\ref{sec:ablation}.

\subsection{The Resistance to Adaptive Attacks}
\label{sec:adaptiveattack}
In this section, we consider adaptive attacks where adversaries understand our watermarking principle but lack knowledge of specific embedding parameters. 
Given that our method modifies character structures for watermark embedding, adversaries may attempt to disrupt extraction by applying perturbations to character geometries. We assess this threat by simulating structure morphing attacks 
\begin{wraptable}{r}{0.48\textwidth}
	\vspace{-1em}
	\centering
	\caption{The robustness of our \name{} against structure morphing attack of varying strengths and font sizes.}
	\label{tab:adaptiveattack}
    \renewcommand{\arraystretch}{1.2} 
	\scalebox{0.75}{
	\begin{tabular}{ccccccc}
	\hline
	Strength &12pt & 16pt & 20pt & 24pt & 28pt & 36pt\\
	\hline
	Light & 97.83 & 98.56 & 98.91 & 99.12 & 99.54 & 99.96\\
	Strong & 96.54 & 97.21 & 97.89 & 98.35 & 98.92 & 99.74 \\
	\hline
	\end{tabular}
	}
	\vspace{-1em}
\end{wraptable}
through elastic deformations of character strokes, which subtly alter stroke contours and angles without compromising semantic identity.
As shown in Table~\ref{tab:adaptiveattack}, \name{} demonstrates robust performance against both light and strong structure morphing attacks across various font sizes. These results confirm that \name{} maintains high extraction accuracy even when adversaries deliberately manipulate character structures, validating its effectiveness in practical adversarial scenarios. In addition, we further assess \name{}’s robustness against additional adaptive scenarios in Appendix~\ref{adaptiveAppendix}, such as common image processing, partial interception, and letter comparison.

\section{Conclusion}
\label{sec:Conclusion}
We presented \name{}, a structure-based document hiding framework that leverages diffusion models to embed data via subtle modifications to character structures. \name{} comprises three key components: adaptive diffusion initialization, guided diffusion encoding, and masked region replacement. Extensive experiments verified \name{}'s superior cross-media robustness, multi-font and multi-language generalizability, high imperceptibility, and resilience against adaptive attacks. We hope our \name{} can provide new insights into document hiding, to facilitate trustworthy sharing and attribution of documents.

\section*{Acknowledgments}
This work is supported by the National Research Foundation, Singapore, Cyber Security Agency of Singapore under its National Cybersecurity R\&D Programme and CyberSG R\&D Cyber Research Programme Office. Any opinions, findings and conclusions or recommendations expressed in these materials are those of the author(s) and do not reflect the views of National Research Foundation, Singapore, Cyber Security Agency of Singapore as well as CyberSG R\&D Programme Office, Singapore.

\bibliographystyle{plain}
\bibliography{main}

\clearpage
\setcounter{page}{1}

\setcounter{section}{0}
\renewcommand{\thesection}{\Alph{section}}

\setcounter{table}{0}
\renewcommand{\thetable}{\Alph{table}}

\setcounter{figure}{0}
\renewcommand{\thefigure}{\Alph{figure}}

\setcounter{equation}{0}
\renewcommand{\theequation}{\Alph{equation}}
\appendix


{\Large \textbf{Appendix}}

\section{Threat Model}
\label{threat}
This paper considers two primary entities: the document owner and an adversary. Following prior works on text watermarking \cite{yang2023autostegafont, yang2023language}, the adversary is assumed to redistribute watermarked documents without authorization through cross-media channels (\eg, print–scan workflows, print–camera capture, and screenshots). The documents may exhibit diverse formats, including multiple font families and sizes, multilingual scripts, handwritten text, and artistic or stylized typography. In general, the document owner aims to ensure that watermark extraction remains robust after such cross-media transmission, thereby enabling reliable provenance tracing and copyright protection in real-world scenarios. We also assume that the adversary may know the high-level design of our method but cannot access the specific embedding parameters used by the document owner, and may attempt to devise adaptive attacks to bypass the watermark.

\section{More Implementation Details}
\label{implementationAppendix}

\vspace{0.3em}
\noindent \textbf{English Dataset.}
The English dataset comprises three fonts: Arial, Times New Roman, and Calibri. For each font at each size, we create 60 text images containing both uppercase and lowercase letters, with character repetition as needed to achieve the required sample size. While image dimensions vary across font sizes, character composition remains consistent. To evaluate each character's data-carrying capacity and robustness, we embed both `0' and `1' bits into the 60 images, generating 120 data samples for extraction accuracy assessment. This procedure is repeated for each font across various sizes using both \name{} and AutoStegaFont methods, with results presented in Tab.~\ref{tab:screenshots}-~\ref{tab:print_camera}.

\vspace{0.3em}
\noindent \textbf{Chinese Dataset.}
The Chinese dataset incorporates two widely adopted fonts: SimSun and SimHei. For each font, we generate the 60 most frequently used Chinese characters using ChatGPT. At each font size, we embed both `0' and `1' bits into these 60 characters, producing 120 data samples for extraction accuracy calculation.

\vspace{0.3em}
\noindent \textbf{Equipment.}
The printer and scanner used in the experiment are HP LaserJet Pro MFP M132nw and Epson V30 SE, both with a resolution of 600 dpi. The shooting device is a Xiaomi 14 mobile phone. 

\section{Additional Details of Adaptive Diffusion Initialization}

\vspace{0.3em}
\noindent\textbf{Module Outputs and Desired Properties.}
Table~\ref{tab:module_outputs} summarizes the outputs and
desired properties of the four modules in adaptive
diffusion initialization.

\begin{table}[h]
    \centering
    \caption{Outputs and desired properties of the four modules
    in adaptive diffusion initialization.}
    \label{tab:module_outputs}
    \small
    \setlength{\tabcolsep}{4pt}
    \renewcommand{\arraystretch}{1.2}
    \begin{tabularx}{\linewidth}{
        @{}p{0.16\linewidth}p{0.25\linewidth}X@{}
    }
    \toprule
    \textbf{Module} & \textbf{Output} &
    \textbf{Desired Properties} \\
    \midrule
    Keypoint \newline Detection & Endpoints and \newline junction points & Accurately represent the glyph skeleton without
    requiring font or character labels \\
    \midrule
    MPE & Handle point $P_h$ and \newline reference point $P_r$
    & $P_h$ should be movable without changing
    character identity, while remaining consistently
    identifiable during extraction \\
    \midrule
    TPE & Target point $P_t$
    & Satisfy the bit-encoding condition while preserving
    the local stroke direction and structural consistency \\
    \midrule
    MDM & Editing mask $M$
    & Cover the required movement region while minimizing
    modifications outside the target structure \\
    \bottomrule
    \end{tabularx}
\end{table}

\begin{figure}[h]
\centering
c\includegraphics[width=0.8\textwidth]{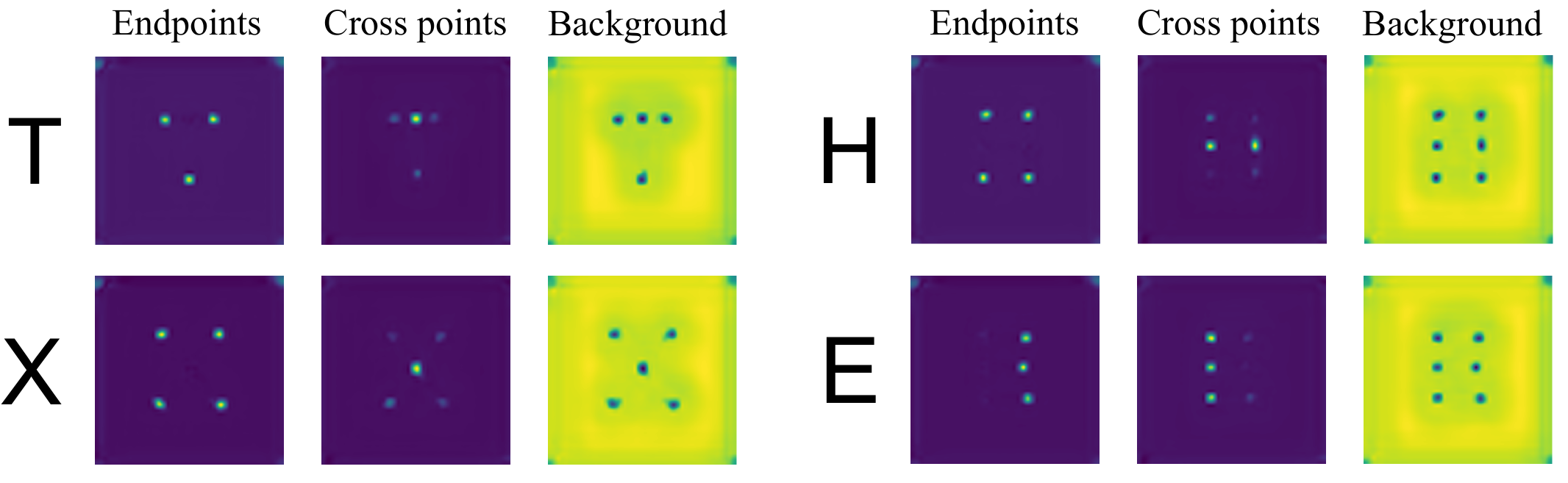}
\caption{Three heatmaps output by the keypoint extraction network, including an endpoint heatmap, a junction point heatmap, and a background heatmap.}
\label{fig:heatmap}
\end{figure}

\vspace{0.3em}
\noindent\textbf{Keypoint Detection.}
Fig.~\ref{fig:heatmap} displays the heatmaps generated by the keypoint extraction network, where pixel values indicate the probability of a location belonging to a specific joint class. In our configuration, the first channel represents endpoints, the second channel represents junction points, and the third channel corresponds to the background.

\begin{algorithm}
	\caption{Movement probability evaluator (MPE).}\label{alg:alg1}
    \label{alg1}
	\textsl{}
	\begin{algorithmic}[1]
		\renewcommand{\algorithmicrequire}{\textbf{Input:}}
		\renewcommand{\algorithmicensure}{\textbf{Output:}}
		\REQUIRE{$ E $, $ C $}
		\ENSURE{$ P_h $ and $ P_r $}
		\FOR{ $ p_i^e \in E $}
		\STATE Generate $ R_i $ for the current $ p_i^e $
		\IF{$\text{len}(R_i) = 0$}
		\STATE // \textit{If $ p_i^e $ has no reference point, the possibility of becoming $ P_h $ is 0.} \\Set initial score $s_{i,j} \gets 0$ for $p_i^e$ 
		\ELSE
		\STATE Set initial score $s_{i,j} \gets 1$ for each $p_{i,j}^r \in R_i$
		\FOR{$j \in \left\{1,...,\text{len}(R_i)\right\}$}
		\STATE // \textit{$ \mathcal{R}^1 $: check the connectivity. }
		\IF{$p_i^e$ and $p_{i,j}^r$ are not on the same stroke}
		\STATE $s_{i,j} \gets s_{i,j} + 1$  
		\ENDIF
		\ENDFOR 
		\STATE // \textit{$ \mathcal{R}^2 $: add one point to the $p_{i,j}^r$ with the smallest $ y_{i,j}^r $ and a score of 2.}
		\\ $j_{\text{min}} = \arg\min_{j} y_{i,j}^r$, $ s_{i,j_{\text{min}}} \gets s_{i,j_{\text{min}}} + 1$
		\STATE // \textit{Choose the $ p^r_i $ with the highest score as the final reference point for the current $ p_i^e $.} \\$ j_{\text{max}} = \arg\max_{j} \, s_{i,j}$ \\ $ \hat{s}_i \gets s_{i,j_{\text{max}}}$, $ \hat{p}_r^i \gets p_{j_{\text{max}}}^r$  
		\ENDIF
		\ENDFOR
		\STATE // \textit{Next, select the final $ P_h $ from $ E $.}
		\STATE $ i_{\text{max}} = \arg\max_{i} \, \hat{s}_i $
		\STATE $ S_{idx} = \{ i \mid \hat{s}_i = \hat{s}_{ i_{\text{max}}} \}$
		\IF{$ \text{len}(S_{\text{idx}}) = 1$ }
		\STATE $ P_h \gets p^e_{i_{\text{max}}} $, $ P_r \gets \hat{p}^r_{i_{\text{max}}} $
		\ELSE 
		\STATE // If there are multiple maximum values, choose the $ p_i^e $ with the smallest $ y-$coordinate as $ P_h $.\\  $ \beta = \arg\min_{i \in S_{\text{idx}}}(p_i^e)_y$
		\\ $ P_h \gets p^e_{\beta} $, $ P_r \gets \hat{p}^r_{\beta} $
		\ENDIF
		\RETURN $ P_h $ and its reference point $ P_r $
	\end{algorithmic}
\end{algorithm}
\section{More Details of MPE}
\label{app-MPE}
\begin{figure}[h]
	\centering
	\includegraphics[width=0.6\textwidth]{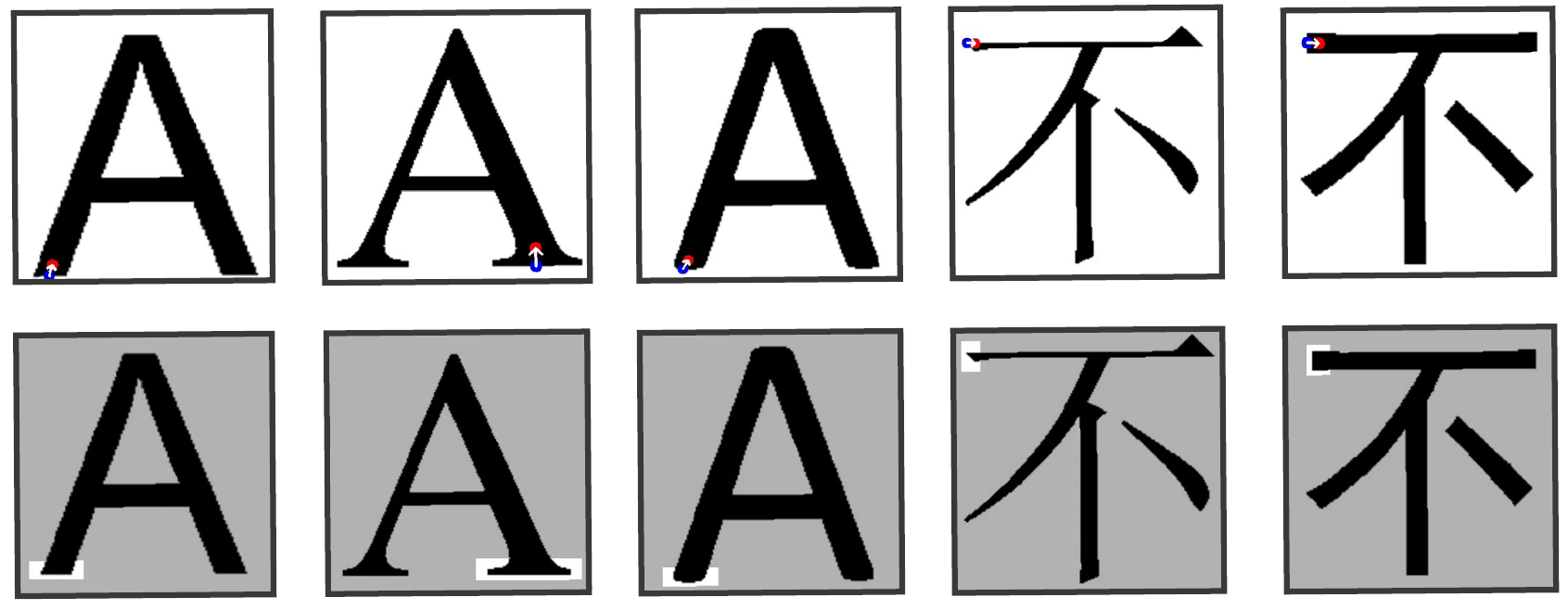}
	\caption{Examples of handle points and target points selected by MPE and TPE, respectively, and masks drawn by MDM.}
	\label{fig:showMDM}
	\vspace{-1em}
	\end{figure}

The proposed Movement Probability Evaluator (MPE) method is designed to automatically determine the handle point for each character, with its main steps discussed in Section~\ref{sec:MPE}. To facilitate a clearer understanding, we provide detailed pseudocode in Algorithm 1, which outlines the workflow and key computational steps of MPE. In addition, we provide some output examples of handle points picked by MPE in Fig.~\ref{fig:showMDM}, where the handle points are marked as blue dots. The red dots are the target points generated by the Target Point Estimation (TPE). For each example, we also present the corresponding masks generated by mask drawing module (MDM), as shown in the second row of Fig.~\ref{fig:showMDM}.

\section{More Ablation Studies}
\label{sec:ablation}

\begin{figure*}[t]
	\centering
	\begin{minipage}[c]{0.48\textwidth} 
	\centering
	\vspace{-0.3em}
	\includegraphics[width=\textwidth]{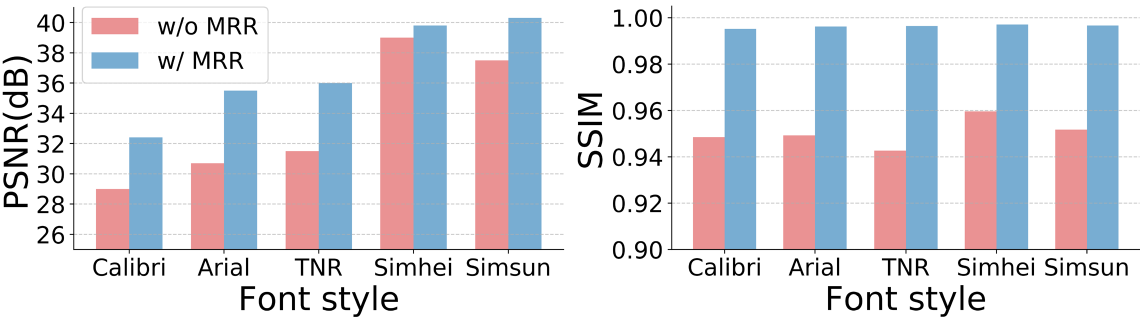}
	\caption{Ablation study on MRR across various fonts in 12pt.}
	\label{fig:mrrablation}
	\end{minipage}
	\hspace{0.2em}
	\begin{minipage}[c]{0.48\textwidth} 
	\centering
	\includegraphics[width=\textwidth]{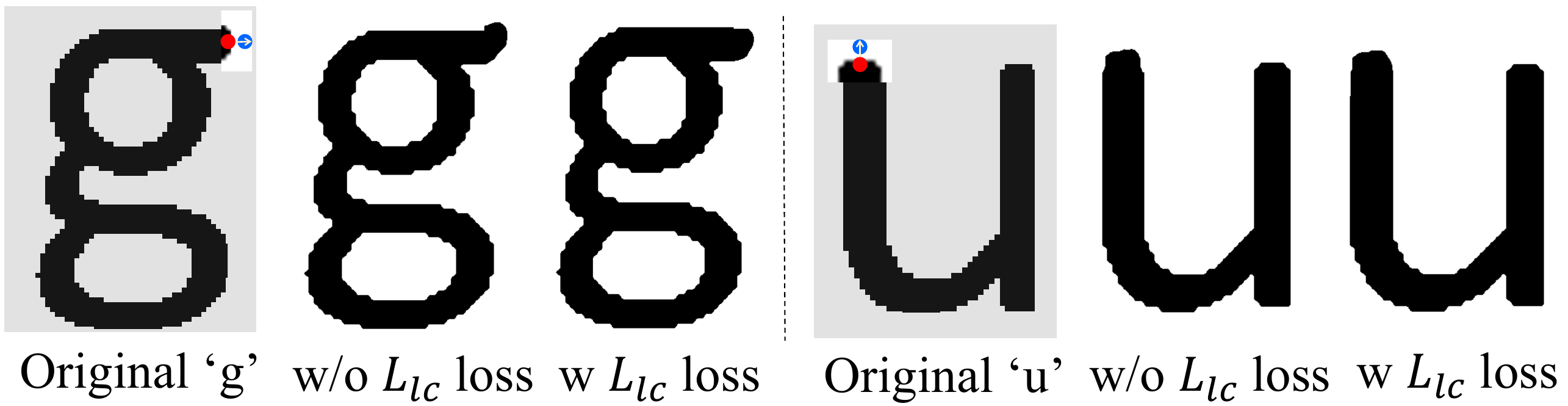} 
	\caption{Ablation study on $ L_{lc} $ loss, taking lowercase letters `g' and `u' in Calibri as examples.}
	\label{fig:lossablation}
	\end{minipage}
	\vspace{-1.5em}
\end{figure*}

\noindent \textbf{Effects of $ L_{lc} $ Loss.}
This loss is designed to enhance consistency of content in the masked area before and after embedding. As shown in Fig.~\ref{fig:lossablation}, the $ L_{lc} $ loss significantly optimizes shape preservation in the masked region, rendering it virtually indistinguishable from the original image.

\vspace{0.3em}
\noindent \textbf{Effects of MRR.}
As shown in Fig.~\ref{fig:mrrablation}, MRR significantly enhances
imperceptibility. Specifically, we observe varying degrees of improvement in PSNR and SSIM across different fonts, which can be attributed to MRR's ability to embed data locally without affecting other regions of the image.


\begin{wraptable}{r}{0.48\textwidth}
\vspace{-1.2em}
\caption{Ablation study on MPE and TPE. (\ding{55}: random point selection operation, \ding{51}: MPE or TPE operation).}
\label{MPETPE}
\scalebox{0.7}{
\begin{tabular}{ccccc}
\toprule
Design & Setting 1 & Setting 2 & Setting 3 & Setting 4 (Ours)\\
\toprule
MPE & \ding{55} & \ding{51} & \ding{55} & \ding{51} \\
TPE	& \ding{55} & \ding{55} & \ding{51} & \ding{51} \\
\midrule
ACC
& 49.95 & 68.75 & 53.50 & \textbf{100} \\
\bottomrule
\end{tabular}
}
\vspace{-1em}
\end{wraptable}
\vspace{0.3em}
\noindent \textbf{Effects of MPE and TPE.}
Recall that MPE aims to select the handle points and TPE determines the target points. To verify their effectiveness, 
we replace them with a random selection and conduct four experiments (\eg, Cases 1-4), as shown in Tab.~\ref{MPETPE}.  
With both MPE and TPE active, extraction is error-free. MPE alone achieves limited accuracy due to probabilistic encoder-decoder synchronization. In the remaining two cases, accuracy approaches 50\%, indicating random decoding.. To further validate the design of MPE, we perform ablation studies on each of its scoring rules, demonstrating their individual necessity. Results are provided in Appendix~\ref{sec:appdixMPEablation}.

\section{Ablation Studies of MPE}
\label{sec:appdixMPEablation}
In Section~\ref{sec: ablation}, we demonstrated that MPE is essential for \name{}. Specifically, MPE selects handle points using three rules. Here, we validate the necessity of each rule through additional ablation studies. The three rules are:
\begin{itemize}
\item \textbf{Rule 1:} Assign each endpoint an initial score of 1.
\item \textbf{Rule 2:} For candidates scoring 1 after Rule 1, increment the score by 1 if the candidate and its reference point lie on different strokes.
\item \textbf{Rule 3:} Among candidates with equal highest scores after Rule 2, the candidate with the smallest y-coordinate is assigned an additional score of 1, and this point is selected as the handle point. 
\end{itemize}

Each rule progressively updates candidate scores, with MPE ultimately selecting the highest-scoring point as the handle point. To validate the necessity of each rule, we design three ablation experiments, each eliminating the effect of one rule on the final scoring:
\begin{itemize}
\item \textbf{w/o Rule 1:} Assign scores to junction points as well rather than restricting the initial scoring to endpoints only.
\item \textbf{w/o Rule 2:} Remove the stroke connectivity constraint, treating all points equally.
\item \textbf{w/o Rule 3:} Eliminate y-coordinate-based selection, randomly choosing among top-scoring candidates when multiple candidates share the highest score.
\end{itemize}

As shown in Table~\ref{tab:mpeablation}, the full version of MPE achieves the best performance across all metrics. The \textbf{w/o Rule 1} experiment demonstrates that without proper initial endpoint scoring, structural disruption occurs, resulting in the lowest PSNR and ACC. The \textbf{w/o Rule 2} experiment reveals that ignoring stroke consistency leads to character invariance, causing embedding failure. Consequently, PSNR increases due to minimal perturbation, while ACC decreases drastically. The \textbf{w/o Rule 3} experiment shows that removing y-coordinate prioritization causes desynchronization between the encoder and decoder, compromising robustness. These findings collectively demonstrate the necessity and effectiveness of each proposed heuristic rule in MPE.

\begin{table}
\centering
\caption{Ablation study on MPE scoring rules, failed cases (\ie, PSNR < 30dB, SSIM < 0.95, and ACC < 85\%) are marked in red.}
\label{tab:mpeablation}
\setlength\tabcolsep{4pt}{
\begin{tabular}{cccc}
\toprule
Method   & PSNR (dB) ↑   & SSIM ↑ & ACC (\%) ↑    \\
\midrule
w/o Rule 1: & \textcolor{red}{25.24} & \textcolor{red}{0.911}  & \textcolor{red}{63.33} \\
w/o Rule 2: & 39.48 & 0.997  & \textcolor{red}{80.67} \\
w/o Rule 3: & 32.82 & 0.965  & \textcolor{red}{68.51} \\
\midrule
MPE (Ours) & 32.10  & 0.962  & 100.00 \\
\bottomrule
\end{tabular}
}
\end{table}

\section{Stage-wise Reliability Analysis}
\label{sec:stage_reliability}

While the end-to-end extraction accuracy reflects the overall robustness of \name{}, it does not isolate errors introduced by individual stages. To provide further insights, we separately evaluate the reliability of character segmentation and keypoint detection.

\noindent\textbf{Character Segmentation.}
We evaluate CRAFT on 50 documents containing mixed-case text, varying font sizes, and dense layouts, achieving a segmentation accuracy of 96.69\%. Segmentation errors do not necessarily lead to extraction failures, as \name{} preserves character spacing and overall glyph
structures. Even merged character regions may remain decodable since our method relies on structural keypoints rather than strict character boundaries.

\noindent\textbf{Keypoint Detection.}
We further evaluate the keypoint detector on 50 manually
annotated characters, achieving a precision of 98.67\%.
Detection errors affect watermark extraction primarily
when they alter the handle or reference points selected
by MPE, or change the final distance-based bit decision.
Minor localization errors that preserve these relationships
may not affect extraction.

\noindent\textbf{Potential Improvements.}
Residual errors could be further mitigated through
error-correcting codes, synchronization markers, or
repeated encoding. These document-level mechanisms
are orthogonal to our core contribution and are therefore
excluded from the current evaluation. They can be
incorporated into the existing framework as a direction
for future work.

\section{Detailed Calculation Process of Mask Drawing Module}
\label{sec:appdixMDM}
After obtaining $\mathcal{M}_c$, which is defined by a four-tuple $ (0, \Gamma_{\text{left}}, H, \Gamma_{\text{right}}-\Gamma_{\text{left}}) $, we hereby illustrate how to calculate $\Gamma_{\text{top}} $ and $ \Gamma_{\text{bottom}} $.
Assume that in $ I_{cover} $, text is black (pixel value 0) and background is white (pixel value 255), we process each column of pixel sequence $ \mathcal{M}_c^i $ of $ \mathcal{M}_c $ by removing its first element to obtain a new sequence $ \widetilde{\mathcal{M}_c^i} $, and appending a large value (\eg, 254) to maintain the same length. We then compute the difference between $ \mathcal{M}_c^i $ and $ \widetilde{\mathcal{M}_c^i} $ to obtain sequence $ \zeta $. The values of 255 in $ \zeta $ mark the transitions from non-black to black pixels. We filter transition points occurring before $ y_h $ and store their indices in $ \zeta_{\text{255}}^i $. Let $ \mathcal{K}_{\text{top}} = \left\{\kappa_{top}^i \middle| i=1,2,...,\Gamma_{\text{right}}-\Gamma_{\text{left}}\right\} $ store the maximum value of $ \zeta_{\text{255}}^i $ for each $ \mathcal{M}_c^i $. If no values in $ \zeta_{\text{255}}^i $ meet the criteria, we set $ \kappa_{top}^i $ to $ y_h $; otherwise, we select the largest value as $ \kappa_{top}^i $. For $ \Gamma_{\text{top}} $, it equals to $\min(y_h, \ y_t, \ \min(\mathcal{K_{\text{top}}}))$.
Similarly, positions with a value of -255 in $ \zeta $ represent the transitions from black to non-black pixels. We select transition points occurring after $ y_h $ and store their indices in $ \zeta_{\text{-255}}^i $. Let $ \mathcal{K}_{\text{bottom}} = \left\{\kappa_{\text{bottom}}^i \middle| i=1,2,...,\Gamma_{\text{right}}-\Gamma_{\text{left}}\right\} $ store the minimum value of $ \zeta_{\text{-255}}^i $ for each $ \mathcal{M}_c^i $. If the length of $ \zeta_{\text{-255}}^i $ is 0, set $ \kappa_{\text{bottom}}^i = y_h $; otherwise, assign the smallest value to $ \kappa_{\text{bottom}}^i $. $ \Gamma_{\text{bottom}} $ is equals to $\max(y_h, \ y_t, \ \max(\mathcal{K_{\text{bottom}}}))$.

\section{Capacity Analysis of \name{}}
\label{sec:capacity}
The embedding capacity of TRACE depends on the number of embeddable characters in a document. To assess its practical capacity, we randomly sample 100 document images from DocLayNet \cite{pfitzmann2022doclaynet} and measure the proportion of characters eligible for watermark embedding. The results show that the proportion
of embeddable characters ranges from 71.96\% to 90.01\%, with an average of 82.38\%.

Since each embeddable character carries one bit of information,
a 10-bit identifier, capable of distinguishing
$2^{10}=1,024$ identities, requires only
$10/0.8238 \approx 13$ original characters on average.
This corresponds to merely a few words or a short sentence,
indicating that typical text-rich documents provide sufficient
capacity for practical watermark embedding.

\section{Efficiency of \name{}}
\label{sec:appdixefficiency}
Given a real image with the resolution of $512 \times 512$, the execution time of different stages in \name{} on a 4090 GPU is as follows: LoRA fine-tuning is around 30 seconds, latent optimization is around 1 minute depending on the magnitude of the hyper-parameter $\mathcal{D}$, the adaptive diffusion initialization is negligible comparing to previous steps (about 0.01 to 0.015 second), and the watermark extraction is extremely fast, taking approximately $0.005$ second.

Arguably, in general, although efficiency is not the main focus of this paper, we hereby discuss several potential strategies to further improve our efficiency. Firstly, consistent with existing approaches \cite{yang2023autostegafont, xiao2018fontcode, qi2019robust}, we can pre-compute a character codebook to mitigate runtime overhead. Under equivalent codebook conditions, our method demonstrates superior robustness and imperceptibility, as evidenced in Tables~\ref{tab:psnrssim}--\ref{tab:print_camera}.
Secondly, our approach leverages DragDiffusion, a point-based image editing method, to subtly adjust character structures. As illustrated in Fig.~\ref{fig:pipeline}, DragDiffusion operates independently from the rest of our pipeline, allowing users to optionally substitute it with more lightweight alternatives, such as InstantDrag \cite{shin2024instantdrag}, to achieve greater efficiency.To demonstrate this flexibility, we replaced DragDiffusion with InstantDrag, which significantly reduces the information embedding time to 0.704 seconds. However, since InstantDrag tends to produce larger structural displacements, invisibility slightly decreases, with PSNR dropping by approximately 0.7 dB and SSIM by 0.0002. Nevertheless, robustness remains largely consistent with DragDiffusion, demonstrating that our framework can flexibly trade off efficiency and imperceptibility without sacrificing watermarking reliability



\section{Additional Generalizability Analysis}

\subsection{Generalizability to Other Languages}
\label{subsec:appdixforeign}
\begin{figure*}
\centering
\includegraphics[width=\textwidth]{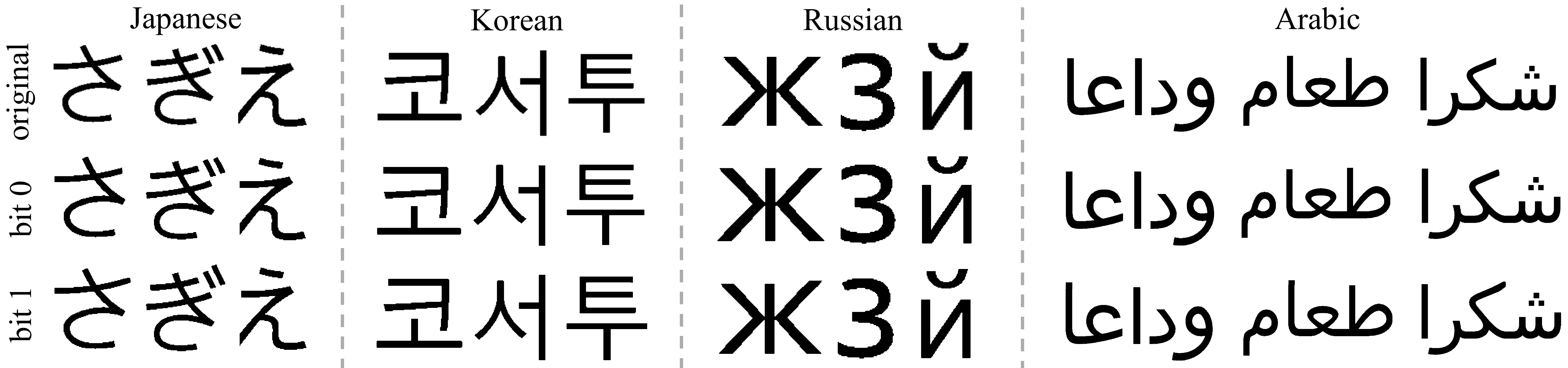}
\caption{Examples of Japanese, Korean, Russian, and Arabic characters embedding '0' and '1' using the \name{} method.}
\label{fig:appdixforeign}
\end{figure*}

In the main manuscript, we evaluated \name{} on two Chinese fonts and three English fonts. To further assess its generalizability, we hereby additionally apply \name{} to fonts from other widely used languages, including Korean, Japanese, Russian, and Arabic. We exclude languages that use alphabetic systems similar to English (e.g., French and Spanish) from this analysis. As shown in Figure ~\ref{fig:appdixforeign}, \name{} successfully embeds distinct bit information into characters across these languages while maintaining good invisibility.

\subsection{Generalizability to Mathematical Formulas}
\label{subsec:formula}
\begin{figure*}
\centering
\includegraphics[width=\textwidth]{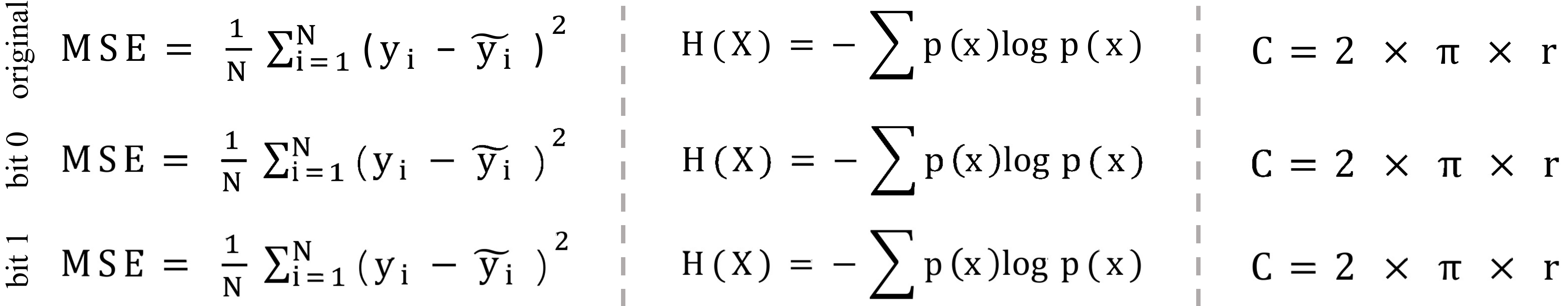}
\caption{Generalizability in mathematical formulas, each character in every formula embeds 0 and 1 bits, respectively.}
\label{fig:appdixformula}
\end{figure*}

Given that mathematical formulas frequently appear in academic and technical documents, we further examine \name{}’s adaptability to this type of content. We select three commonly used formulas as examples and embed both 0 and 1 bits into their constituent characters, as illustrated in Figure~\ref{fig:appdixformula}. \name{} consistently embeds information across these formulas while maintaining strong visual invisibility.

\section{The Resistance to More Adaptive Attacks}
\label{adaptiveAppendix}

\subsection{Resistance to Common Image Processing}
\label{imageprocessApp}
If attackers lacks knowledge of our method's design, they may eliminate the embedded information through common image processing. To evaluate resistance to such attacks, we apply Gaussian noise with a variance of 0.03 and a 3×3 Gaussian blur to 30 encoded character images in 12pt Calibri font. We then extract the embedded information from the attacked images and calculate the accuracy (ACC). The results show that the watermark extraction accuracy remains at 100\% under both attacks, confirming that our method effectively withstands common image processing attacks.

\subsection{Resistance to Letter Comparing}
\label{app-sec-lettercompare}
We admit that, due to the inherently sparse pixel distribution in text images, text-based watermarking methods may face a fundamental challenge: direct comparison between the encoded and original images may reveal watermark presence or leak embedded content. For instance, in font-based approaches employing fixed character codebooks, attackers could potentially infer embedded information through character-to-codebook matching.
However, in most practical scenarios, the primary goal of an attacker is to forge authentication data to pass document validation tests \cite{wu2004data}. To mitigate this threat, we employ a 128-bit secret key for message encryption prior to embedding. The cryptographic security of this approach is evident: the probability of successfully compromising a 128-bit key is $2^{-128}$, which is computationally infeasible. Consequently, even if an attacker manages to extract the ciphertext from the image, extracts the ciphertext from the image, they cannot decrypt the underlying plaintext or forge valid authentication information.

\subsection{Resistance to Partial Interception}
Attackers may only capture the content they are interested in.  In this section, we evaluate \name{}'s resistance to such partial interception attacks. Typically, minimal data, such as an ID or name, is sufficient to indicate identity. \name{} embeds one bit per character, enabling repeated data hiding to resist partial interception. To demonstrate this capability, we evaluate the impact of both embedded bits and the number of consecutively selected words on average extraction accuracy. Tab.~\ref{partialselection} shows the average accuracy over 10,000 extractions. As seen, with an appropriate amount of embedded information, repeated embedding ensures robustness. It is worth noting that attackers are unlikely to extract only a few words, as it would be meaningless.

\begin{table}[h]
    \centering
    \caption{Impact of embedded bits and consecutive word count on average extraction accuracy.}
    \label{partialselection}
    \begin{tabular}{ccccc}
        \hline
        Number of words & 5 & 10 & 15 & 20 \\
        \hline
        8bit  & 97.78 & 100.00 & 100.00 & 100.00 \\
        16bit & 59.76 & 99.62 & 100.00 & 100.00 \\
        32bit & 0.37  & 61.16 & 99.03  & 100.00 \\
        \hline
    \end{tabular}
\end{table}

\section{Potential Limitations and Future Directions}
\label{limitation}
Firstly, \name{} primarily embeds information through the endpoints of character structures. For characters without endpoints (\eg, `o'), information cannot be embedded. However, we argue that this limitation is mild, as such characters constitute only a small portion of the document, and we can always embed information using other characters. To further improve generalizability, we will design specific embedding rules for characters without keypoints in the future. 

Secondly, while LoRA fine-tuning in DragDiffusion \cite{shi2024dragdiffusion} introduces additional time overhead, we have mitigated this limitation in Section~\ref{sec:appdixefficiency} by demonstrating that DragDiffusion can be substituted with lightweight alternatives such as InstantDrag, significantly reducing embedding time to 0.704 seconds. Further efficiency optimizations remain a direction for future work.

Thirdly, the embedding capacity of TRACE is constrained by the number of embeddable characters. Extremely short text, such as titles, identifiers, or isolated fields, may not contain sufficient characters to accommodate the desired watermark payload. Nevertheless, TRACE primarily targets typical documents containing a reasonable amount of text, for which sufficient embedding capacity is generally available, as demonstrated in our capacity analysis.

\section{Societal Impacts}
\label{app-social}
In this paper, we focus on the secure application and protection of document hiding techniques. Specifically, we reveal that existing document watermarking methods struggle to balance robustness, generalizability, and invisibility. We propose \name, a character structure-based encoding method that achieves strong robustness, wide generalizability, and high invisibility simultaneously. Our approach enables creators and organizations to effectively track document distribution paths and prevent unauthorized dissemination. This has significant value for protecting intellectual property and preventing confidential information leakage. This work raises no general ethical concerns, as our method is purely designed for digital document copyright protection and unauthorized distribution tracking, without introducing any potentially misusable new capabilities. We believe this technology will provide necessary protection tools for content creators and organizations without posing additional risks to information security.

\section{Discussions of OCR Attacks}

Following prior works \cite{yang2023language, yang2023autostegafont, xiao2018fontcode}, this paper does not consider OCR-based attacks, as OCR fundamentally transforms the watermark carrier itself. Once the document is converted through OCR, the original digital carrier ceases to exist, making subsequent evaluation of watermark performance largely meaningless. In contrast, this paper aims to protect the document carrier itself rather than only the textual content.

Furthermore, OCR attacks are not universally applicable, as they fail in two important scenarios. First, OCR systems cannot faithfully reproduce stylized fonts such as artistic or handwriting styles. Our method is, to the best of our knowledge, the first document watermarking scheme that generalizes to these font types, making it effective in settings where OCR-based attacks are inherently limited. Second, the authentication of officially sealed documents fundamentally depends on the physical presence of the seal. Even if OCR successfully extracts the textual content, the document cannot be verified without the original seal, rendering such attacks invalid in practice.

To the best of our knowledge, the only text watermarking techniques capable of withstanding OCR-based attacks are linguistic approaches that alter the original textual content. However, such alterations are typically infeasible in formal documents where content integrity is mandatory. Besides, as observed in ~\cite{yang2023autostegafont}, existing linguistic watermarking schemes cannot guarantee that the modified text remains fully semantically equivalent to the original.

\section{Extended Qualitative Analysis}
\label{sec:appdixrelatedwork}
\subsection{Visual Comparison to Baselines}
\label{compareAppendix}

\begin{figure*}
\centering
\includegraphics[width=\linewidth]{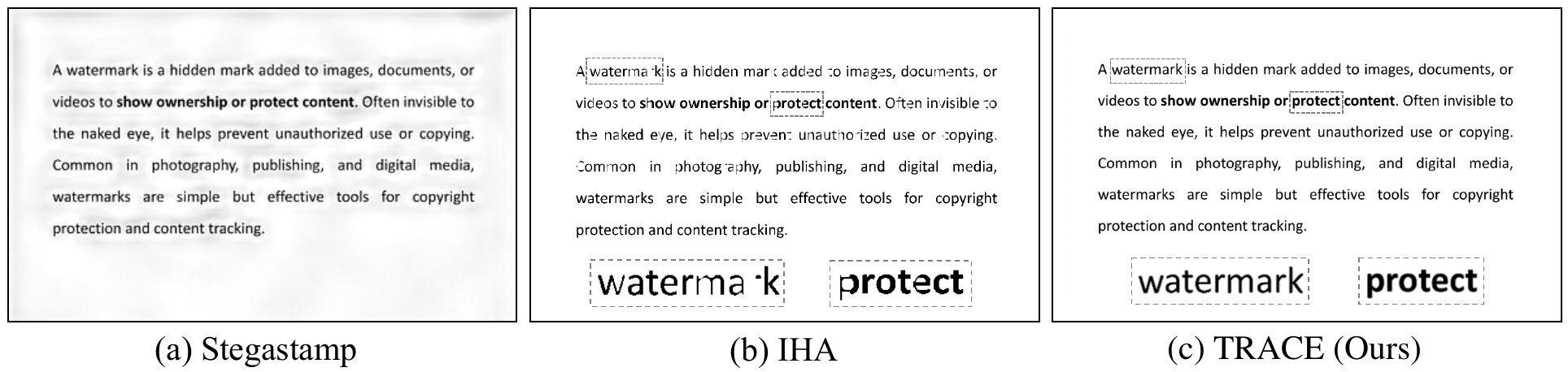}
\caption{The visual effects of our method and the comparison methods.}
\label{fig:documentcompare}
\end{figure*}

We conduct comparisons with the image watermarking method StegaStamp \cite{tancik2020stegastamp} and the information hiding approach (IHA) \cite{yang2023language} in Section~\ref{sec:Experiment} of the main text. Fig.~\ref{fig:documentcompare} presents a visual quality comparison of these document watermarking methods. As shown, StegaStamp introduces noticeable artifacts after information embedding, while IHA leaves overly visible traces that alter the original appearance of certain characters. In contrast, our method achieves superior performance by preserving character appearance with minimal perceptible changes.

\subsection{Comparative Analysis of Existing Methods}
In the main experimental section, we conduct quantitative comparisons with three representative methods \cite{yang2023autostegafont, tancik2020stegastamp, yang2023language}. However, other document watermarking methods are excluded from quantitative analysis due to their inherent limitations and instead presented in a broader qualitative comparison in Tab.~\ref{tab:comparerelatedwork}. Part of the discussion in this section is inspired by the analysis in \cite{yang2023language}.

Specifically, as representative of font-based methods,  \cite{qi2019robust, xiao2018fontcode} demonstrate robustness against cross-media transmission (\eg, print-scan or screenshot). However, they suffer from limited generalizability. In particular, the diversity of handwritten and artistic fonts makes it difficult for font-based encoding schemes to handle all scenarios. Besides, these methods often fail when encountering unseen languages or formulaic symbols, as they typically rely on manually crafted character codebooks that are not easily extensible.

For image-based methods \cite{li2023robust, huang2024robust}, information is embedded by inverting black and white pixels. As a result, they lack generalizability to artistic fonts, which are typically colorful and thus incompatible with such binary manipulations.

Format-based method \cite{brassil1999copyright} rely on layout-level manipulations. While they can be effective in controlled environments, camera-based attacks often introduce distortions (\eg, altered spacing), which undermine the stability of embedding features and significantly affect robustness.

Linguistic-based approaches \cite{kirchenbauer2023watermark, dathathri2024scalable, yang2022tracing, he2022cater} embed information by modifying syntactic or semantic properties of the text. However, due to structural and semantic constraints, such substitutions are not supported in all languages. Moreover, these modifications typically alter the original content, thereby compromising semantic integrity and limiting their applicability in scenarios that require strict preservation of textual meaning, such as legal documents, medical records, or academic publications.

\begin{table*}
\centering
\caption{Qualitative comparison with related work from different perspectives.}
\label{tab:comparerelatedwork}
\vspace{-0.5em}
\footnotesize
\scalebox{0.8}{
\begin{tabular}{c@{\hspace{6pt}}c@{\hspace{6pt}}ccc@{\hspace{6pt}}cccc@{\hspace{6pt}}c@{\hspace{6pt}}}
\hline
\multirow{2}{*}{\raisebox{-0.6ex}[0pt][0pt]{\textbf{Category}}} & \multirow{2}{*}{\raisebox{-0.6ex}[0pt][0pt]{\textbf{Method}}} & \multicolumn{3}{c}{\textbf{Robustness}} &\multicolumn{4}{c}{\textbf{Generalizability}}&\multirow{2}{*}{\raisebox{-0.6ex}[0pt][0pt]{\textbf{Integrity}}}\\
\cmidrule(lr){3-5}\cmidrule(lr){6-9}
&& Print-scan & Print-camera & Screenshot & Handwritten & artistic & formula & Language-universality &  \\
\hline
Font-based &\cite{qi2019robust, xiao2018fontcode}&\ding{51}&\ding{51}&\ding{51}&\ding{55}&\ding{55}&\ding{55}&\ding{55}&\ding{51}\\
Image-based &\cite{li2023robust, huang2024robust}&\ding{51}&\ding{51}&\ding{51}&\ding{51}&\ding{55}&\ding{51}&\ding{51}&\ding{51} \\
Format-based &\cite{brassil1999copyright} &\ding{51}&\ding{55}&\ding{51}&\ding{51}&\ding{51}&\ding{51}&\ding{51}&\ding{51}\\
Linguistic-based &\cite{kirchenbauer2023watermark, dathathri2024scalable, yang2022tracing, he2022cater} & \ding{51}&\ding{51}&\ding{51}&\ding{55}&\ding{51}&\ding{55}&\ding{55}&\ding{55}\\
\multicolumn{2}{c}{\textbf{\name{} (Ours)}}&\ding{51}&\ding{51}&\ding{51}&\ding{51}&\ding{51}&\ding{51}&\ding{51}&\ding{51}\\
\hline
\end{tabular}
}
\end{table*}
\section{More Visual Cases}
\begin{figure*}[h]
	\centering
	\includegraphics[width=0.95\textwidth]{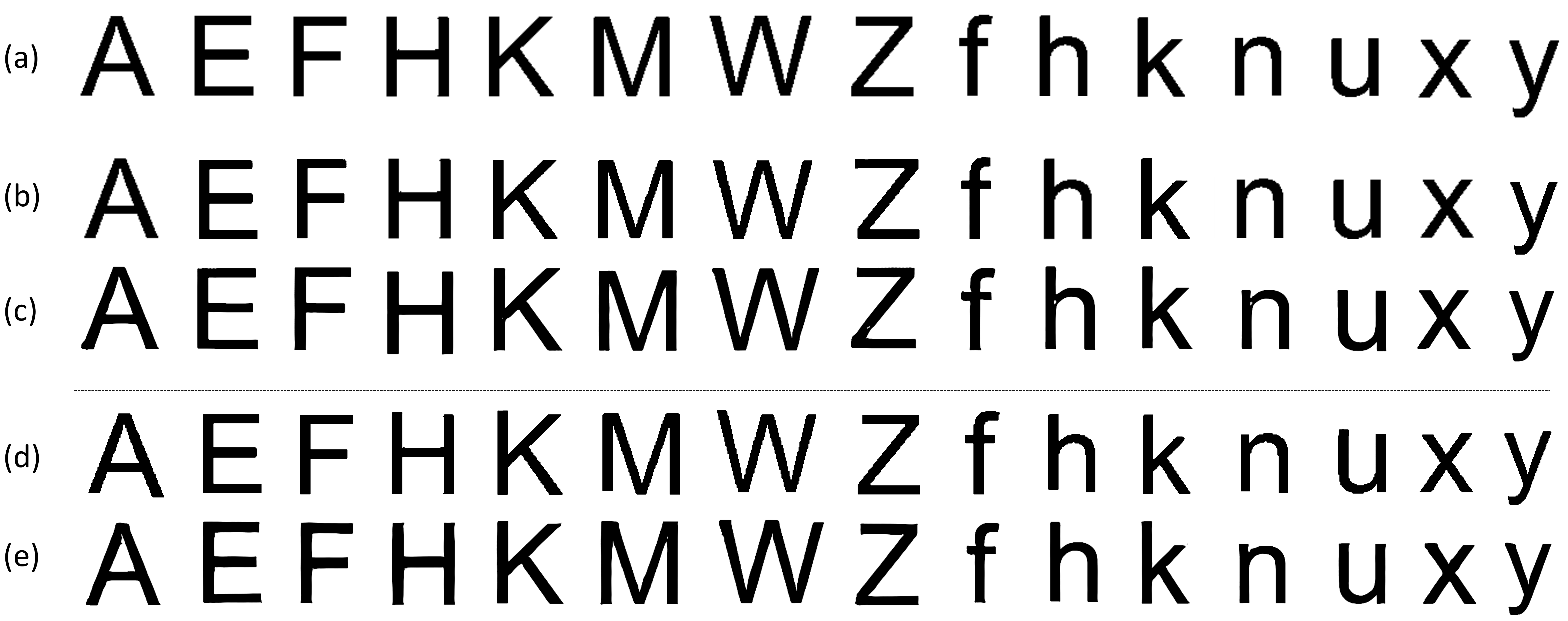}
	\caption{Visual comparisons of our \name{} and the comparison method AutoStegaFont \cite{yang2023autostegafont} on encoded images of the Arial font, (a) original images, (b) encoded images embedding “0” by \name{}, (c) encoded images embedding “1” by AutoStegaFont, (d) encoded images embedding “1” by \name{}, (e)encoded images embedding “0” by AutoStegaFont.}
	\label{arial15}
\end{figure*}

\begin{figure*}
	\centering
	\includegraphics[width=0.95\textwidth]{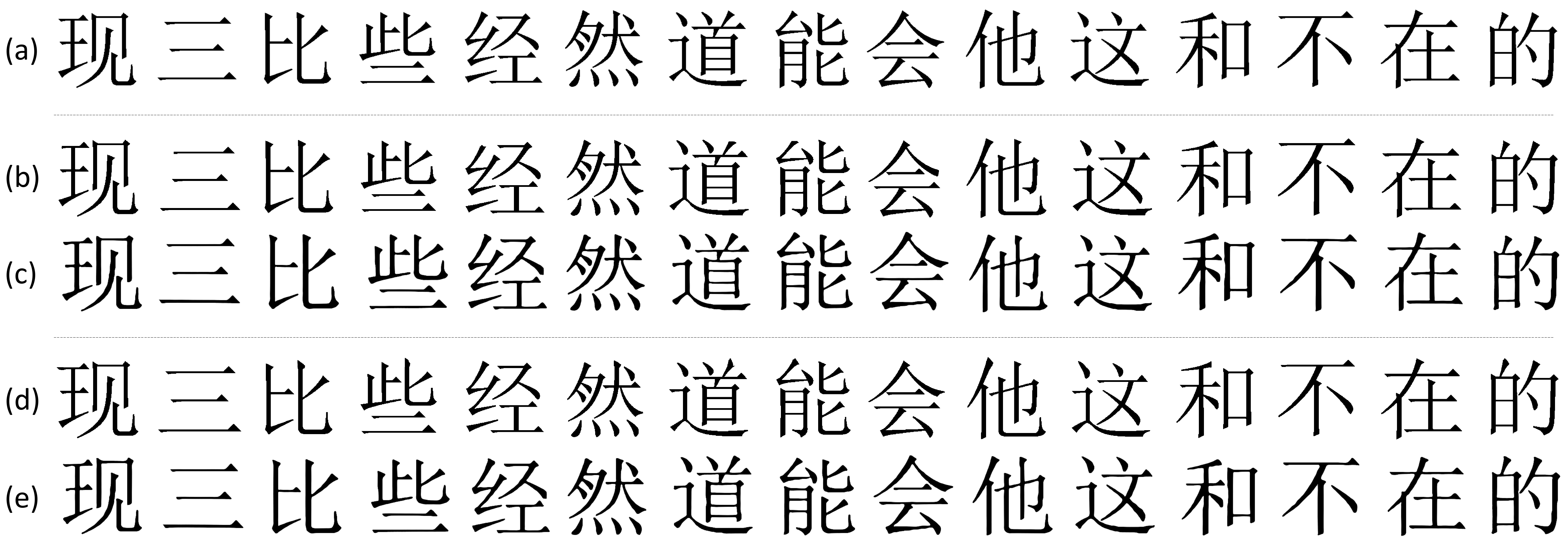}
	\caption{Visual comparisons of our \name{} and the comparison method AutoStegaFont \cite{yang2023autostegafont} on encoded images of the Simsun font, (a) original images, (b) encoded images embedding “0” by \name{}, (c) encoded images embedding “1” by AutoStegaFont, (d) encoded images embedding “1” by \name{}, (e)encoded images embedding “0” by AutoStegaFont.}
	\label{simsun15}
\end{figure*}

In this section, we make more qualitative comparisons between \name{} and AutoStegaFont in Fig.~\ref{arial15}-\ref{simsun15}. The results consistently show that our method achieves significantly better visual quality than AutoStegaFont. Furthermore, it is important to note that AutoStegaFont requires pre-training on the target font before encoding its characters, which introduces considerable complexity and inconvenience in practical applications.
\begin{figure*}[h]
	\centering
	\includegraphics[width=0.95\textwidth]{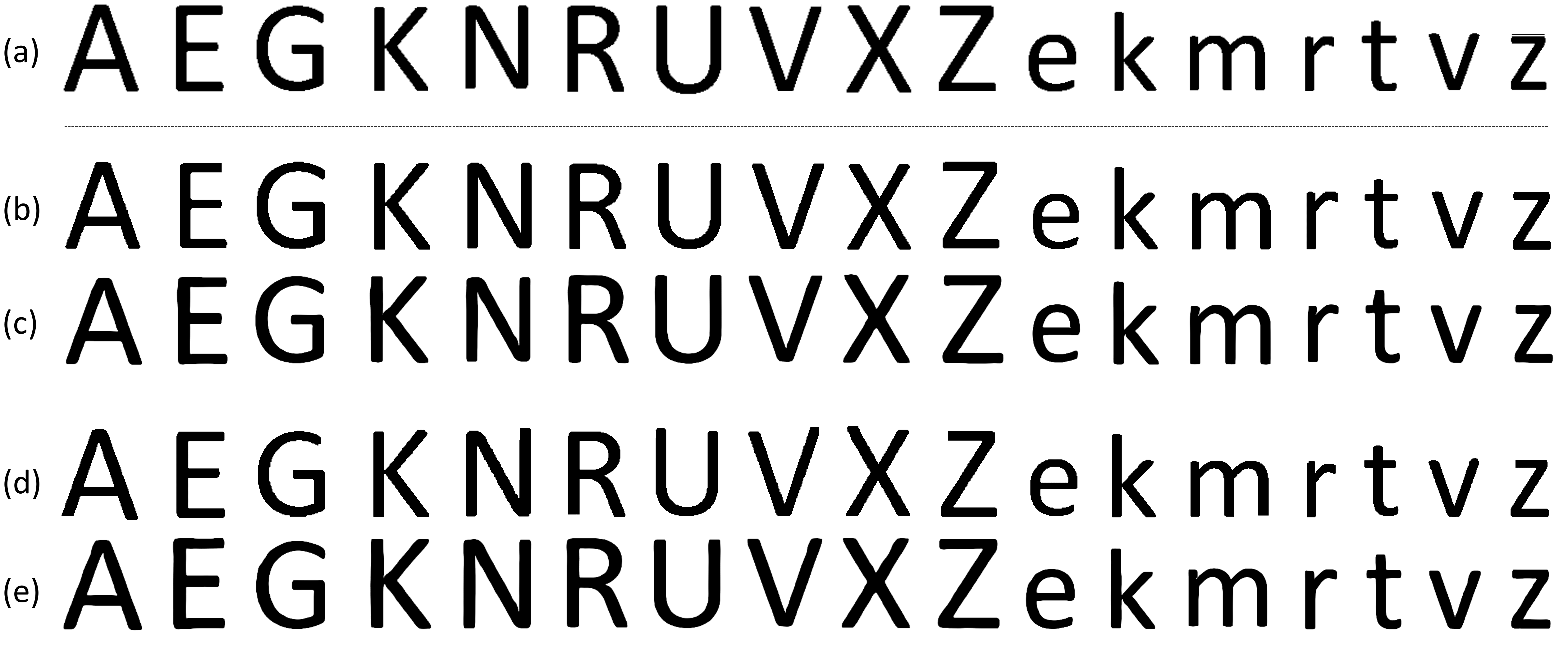}
	\caption{Visual comparisons of our \name{} and the comparison method AutoStegaFont \cite{yang2023autostegafont} on encoded images of the Calibri font, (a) original images, (b) encoded images embedding “0” by \name{}, (c) encoded images embedding “1” by AutoStegaFont, (d) encoded images embedding “1” by \name{}, (e)encoded images embedding “0” by AutoStegaFont.}
\end{figure*}
\begin{figure*}[h]
	\centering
	\includegraphics[width=0.95\textwidth]{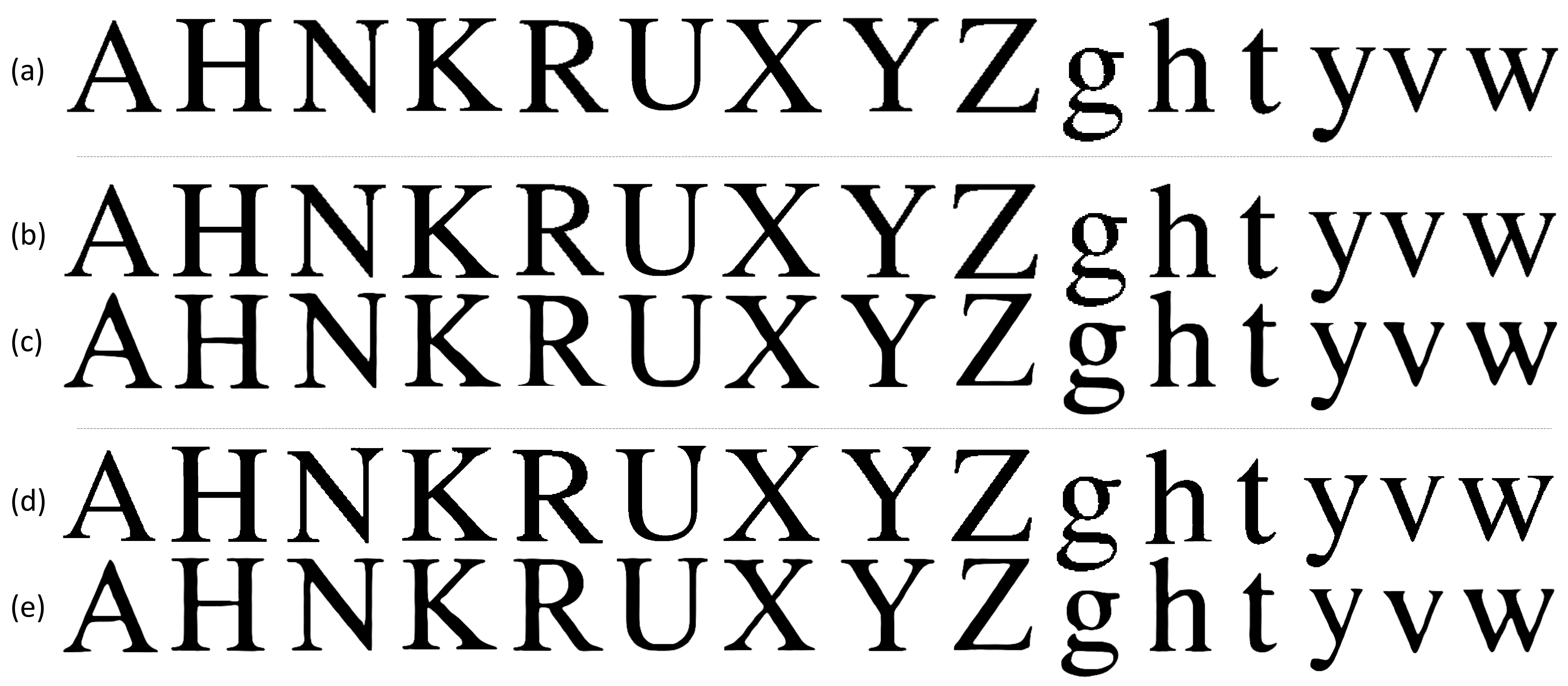}
	\caption{Visual comparisons of our \name{} and the comparison method AutoStegaFont \cite{yang2023autostegafont} on encoded images of the Times New Roman font, (a) original images, (b) encoded images embedding “0” by \name{}, (c) encoded images embedding “1” by AutoStegaFont, (d) encoded images embedding “1” by \name{}, (e)encoded images embedding “0” by AutoStegaFont.}
\end{figure*}
\begin{figure*}[h]
	\centering
	\includegraphics[width=0.95\textwidth]{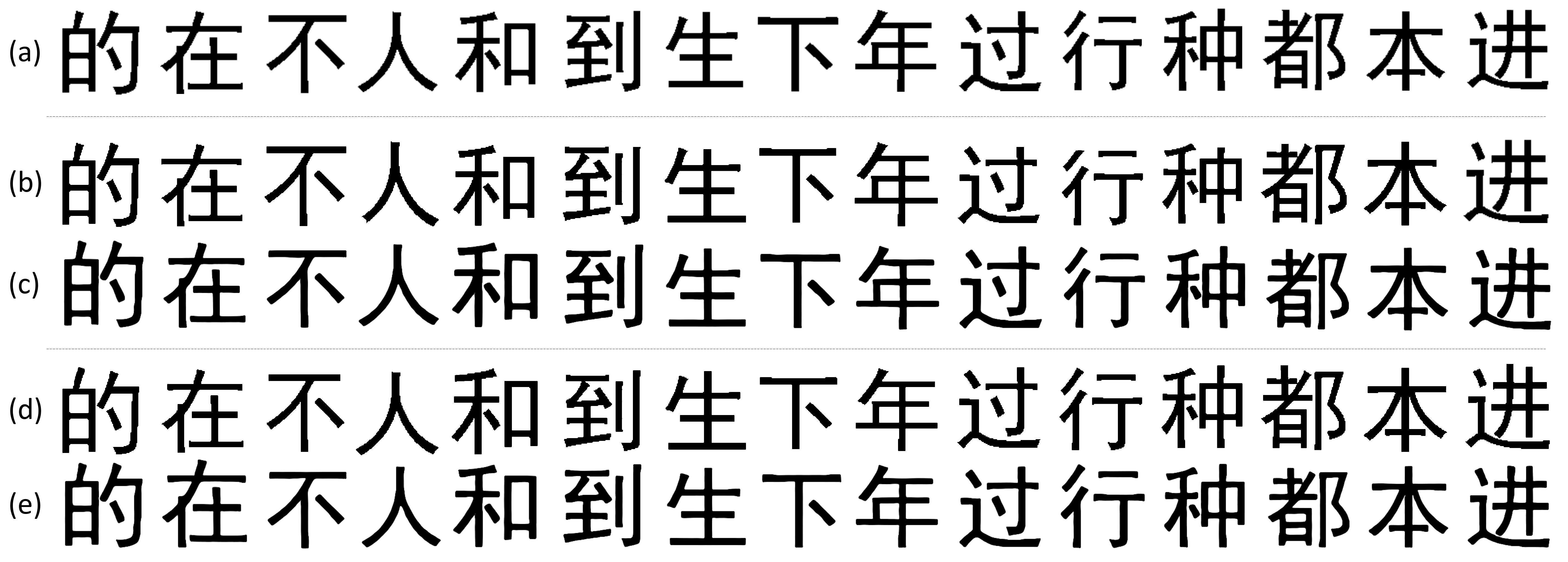}
	\caption{Visual comparisons of our \name{} and the comparison method AutoStegaFont \cite{yang2023autostegafont} on encoded images of the Simhei font, (a) original images, (b) encoded images embedding “0” by \name{}, (c) encoded images embedding “1” by AutoStegaFont, (d) encoded images embedding “1” by \name{}, (e)encoded images embedding “0” by AutoStegaFont.}
\end{figure*}

\clearpage
\input{checklist.tex}

\end{document}

%% file: checklist.tex
\section*{NeurIPS Paper Checklist}

The checklist is designed to encourage best practices for responsible machine learning research, addressing issues of reproducibility, transparency, research ethics, and societal impact. Do not remove the checklist: {\bf The papers not including the checklist will be desk rejected.} The checklist should follow the references and follow the (optional) supplemental material.  The checklist does NOT count towards the page
limit. 

Please read the checklist guidelines carefully for information on how to answer these questions. For each question in the checklist:
\begin{itemize}
    \item You should answer \answerYes{}, \answerNo{}, or \answerNA{}.
    \item \answerNA{} means either that the question is Not Applicable for that particular paper or the relevant information is Not Available.
    \item Please provide a short (1--2 sentence) justification right after your answer (even for \answerNA). 
\end{itemize}

{\bf The checklist answers are an integral part of your paper submission.} They are visible to the reviewers, area chairs, senior area chairs, and ethics reviewers. You will also be asked to include it (after eventual revisions) with the final version of your paper, and its final version will be published with the paper.

The reviewers of your paper will be asked to use the checklist as one of the factors in their evaluation. While \answerYes{} is generally preferable to \answerNo{}, it is perfectly acceptable to answer \answerNo{} provided a proper justification is given (e.g., error bars are not reported because it would be too computationally expensive'' or ``we were unable to find the license for the dataset we used''). In general, answering \answerNo{} or \answerNA{} is not grounds for rejection. While the questions are phrased in a binary way, we acknowledge that the true answer is often more nuanced, so please just use your best judgment and write a justification to elaborate. All supporting evidence can appear either in the main paper or the supplemental material, provided in appendix. If you answer \answerYes{} to a question, in the justification please point to the section(s) where related material for the question can be found.

IMPORTANT, please:
\begin{itemize}
    \item {\bf Delete this instruction block, but keep the section heading ``NeurIPS Paper Checklist"},
    \item  {\bf Keep the checklist subsection headings, questions/answers and guidelines below.}
    \item {\bf Do not modify the questions and only use the provided macros for your answers}.
\end{itemize}


\begin{enumerate}

\item {\bf Claims}
    \item[] Question: Do the main claims made in the abstract and introduction accurately reflect the paper's contributions and scope?
    \item[] Answer: \answerTODO{} 
    \item[] Justification: \justificationTODO{}

\item {\bf Limitations}
    \item[] Question: Does the paper discuss the limitations of the work performed by the authors?
    \item[] Answer: \answerTODO{} 
    \item[] Justification: \justificationTODO{}

\item {\bf Theory assumptions and proofs}
    \item[] Question: For each theoretical result, does the paper provide the full set of assumptions and a complete (and correct) proof?
    \item[] Answer: \answerTODO{} 
    \item[] Justification: \justificationTODO{}

    \item {\bf Experimental result reproducibility}
    \item[] Question: Does the paper fully disclose all the information needed to reproduce the main experimental results of the paper to the extent that it affects the main claims and/or conclusions of the paper (regardless of whether the code and data are provided or not)?
    \item[] Answer: \answerTODO{} 
    \item[] Justification: \justificationTODO{}

\item {\bf Open access to data and code}
    \item[] Question: Does the paper provide open access to the data and code, with sufficient instructions to faithfully reproduce the main experimental results, as described in supplemental material?
    \item[] Answer: \answerTODO{} 
    \item[] Justification: \justificationTODO{}

\item {\bf Experimental setting/details}
    \item[] Question: Does the paper specify all the training and test details (e.g., data splits, hyperparameters, how they were chosen, type of optimizer) necessary to understand the results?
    \item[] Answer: \answerTODO{} 
    \item[] Justification: \justificationTODO{}

\item {\bf Experiment statistical significance}
    \item[] Question: Does the paper report error bars suitably and correctly defined or other appropriate information about the statistical significance of the experiments?
    \item[] Answer: \answerTODO{} 
    \item[] Justification: \justificationTODO{}

\item {\bf Experiments compute resources}
    \item[] Question: For each experiment, does the paper provide sufficient information on the computer resources (type of compute workers, memory, time of execution) needed to reproduce the experiments?
    \item[] Answer: \answerTODO{} 
    \item[] Justification: \justificationTODO{}
    
\item {\bf Code of ethics}
    \item[] Question: Does the research conducted in the paper conform, in every respect, with the NeurIPS Code of Ethics \url{https://neurips.cc/public/EthicsGuidelines}?
    \item[] Answer: \answerTODO{} 
    \item[] Justification: \justificationTODO{}

\item {\bf Broader impacts}
    \item[] Question: Does the paper discuss both potential positive societal impacts and negative societal impacts of the work performed?
    \item[] Answer: \answerTODO{} 
    \item[] Justification: \justificationTODO{}
    
\item {\bf Safeguards}
    \item[] Question: Does the paper describe safeguards that have been put in place for responsible release of data or models that have a high risk for misuse (e.g., pre-trained language models, image generators, or scraped datasets)?
    \item[] Answer: \answerTODO{} 
    \item[] Justification: \justificationTODO{}

\item {\bf Licenses for existing assets}
    \item[] Question: Are the creators or original owners of assets (e.g., code, data, models), used in the paper, properly credited and are the license and terms of use explicitly mentioned and properly respected?
    \item[] Answer: \answerTODO{} 
    \item[] Justification: \justificationTODO{}

\item {\bf New assets}
    \item[] Question: Are new assets introduced in the paper well documented and is the documentation provided alongside the assets?
    \item[] Answer: \answerTODO{} 
    \item[] Justification: \justificationTODO{}

\item {\bf Crowdsourcing and research with human subjects}
    \item[] Question: For crowdsourcing experiments and research with human subjects, does the paper include the full text of instructions given to participants and screenshots, if applicable, as well as details about compensation (if any)? 
    \item[] Answer: \answerTODO{} 
    \item[] Justification: \justificationTODO{}

\item {\bf Institutional review board (IRB) approvals or equivalent for research with human subjects}
    \item[] Question: Does the paper describe potential risks incurred by study participants, whether such risks were disclosed to the subjects, and whether Institutional Review Board (IRB) approvals (or an equivalent approval/review based on the requirements of your country or institution) were obtained?
    \item[] Answer: \answerTODO{} 
    \item[] Justification: \justificationTODO{}

\item {\bf Declaration of LLM usage}
    \item[] Question: Does the paper describe the usage of LLMs if it is an important, original, or non-standard component of the core methods in this research? Note that if the LLM is used only for writing, editing, or formatting purposes and does \emph{not} impact the core methodology, scientific rigor, or originality of the research, declaration is not required.
    \item[] Answer: \answerTODO{} 
    \item[] Justification: \justificationTODO{}

\end{enumerate}